\documentclass{article} 
\usepackage{graphicx}

\usepackage[utf8]{inputenc}
\usepackage{amsmath}
\usepackage{amssymb}
\usepackage[a4paper, margin=2.5cm]{geometry}
\usepackage{parskip}
\usepackage[font=footnotesize, labelfont=bf]{caption}
\usepackage[utf8]{inputenc}
\usepackage{authblk}
\usepackage{xcolor} 

\usepackage{amsmath,amsfonts,bm}

\def\eqref#1{equation~\ref{#1}}

\def\1{\bm{1}}

\DeclareMathAlphabet{\mathsfit}{\encodingdefault}{\sfdefault}{m}{sl}
\SetMathAlphabet{\mathsfit}{bold}{\encodingdefault}{\sfdefault}{bx}{n}

\usepackage{hyperref}
\usepackage{url}

\usepackage{makecell}

\title{MODIS: Multi-Omics Data Integration for Small and unpaired datasets}

\author[1]{Daniel Lepe-Soltero}
\author[2]{Thierry Artières}
\author[1]{Anaïs Baudot}
\author[1*]{Paul Villoutreix}
\affil[1]{Aix Marseille Univ, INSERM, MMG, Marseille, France}
\affil[2]{Aix Marseille Univ \& Ecole Centrale de Marseille, CNRS, LIS, Marseille, France}
\affil[*]{Correspondence: Paul Villoutreix,  \href{mailto:paul.villoutreix@univ-amu.fr}{paul.villoutreix@univ-amu.fr}}

\newcommand{\verysmall}{\fontsize{7}{8.4}\selectfont}

\begin{document}
\maketitle

\begin{abstract}
An important objective in computational biology is the efficient handling of multi-omics data. These heterogeneous multimodal datasets provide complementary information about biological systems and their integration into a joint representation is expected to improve the characterization of the underlying processes. The task of integration comes however with challenges: multi-omics data are most often unpaired (requiring diagonal integration), partially labeled with information about biological conditions, and in some situations such as rare diseases, only very small datasets are available. We present MODIS, which stands for Multi-Omics Data Integration for Small and unpaired datasets, a semi-supervised framework designed to account for these particular challenges. To address the challenge of very small datasets, we propose to exploit the information contained in larger multi-omics databases by training our model on a large reference database and a small target dataset simultaneously, effectively turning the challenge into a problem of learning with class imbalance. MODIS performs diagonal integration on unpaired samples by learning a probabilistic coupling of the heterogeneous data modalities into a shared latent space, leveraging class-labels to align modalities despite class imbalance and data scarcity. The architecture combines multiple variational auto-encoders, a class classifier and an adversarially trained modality classifier. To ensure training stability, we adapted a regularized relativistic GAN loss to this setting. We first validate MODIS on a synthetic dataset to assess the level of supervision needed for accurate alignment and to quantify the impact of class imbalance on predictive performance. We then apply our approach to the large public TCGA database, considering between 10 and 34 classes (cancer types and normal tissue). MODIS demonstrates high prediction accuracy, robust performance with limited supervision, and stability to class imbalance. These results position MODIS as a promising solution for challenging integration scenarios, particularly diagonal integration with a small number of samples, typical of rare diseases studies. The code for MODIS is available at \href{https://github.com/VILLOUTREIXLab/MODIS}{https://github.com/VILLOUTREIXLab/MODIS}. 

\end{abstract}

\section{Introduction}
The recent explosion in multi-modal and particularly multi-omics data holds immense promise for understanding complex biological mechanisms and ultimately to help address disease-related challenges. Rare diseases in particular could benefit from multi-omics approaches. These diseases are defined as diseases affecting less than one person in 2000, are characterized by diagnostic deadlocks, poor understanding of disease pathophysiological mechanisms, and few therapeutic options,  \cite{banerjee2023machine}. To take full advantage of these multi-omics datasets, it is useful to derive a joint representation of the heterogeneous data types in a lower dimensional space, a task named multi-omics data integration. This joint representation helps disease identification by classifying or clustering samples, leading subsequently to biomarker discovery \cite{cantini2021benchmarking, hirst2025motl}. Moreover, since it is not always possible to acquire samples in multiple omics simultaneously, it has been proposed to use the joint representation as a basis for generating missing modalities \cite{yang2021multi}. Various strategies have been implemented for multi-omics integration, ranging from vertical approaches that combine layers of omics data for paired samples to diagonal and mosaic strategies that enable the integration of diverse combinations of unpaired and paired samples across datasets \cite{argelaguet2021computational}. We are interested here in the applicability of data integration for disease identification and subtyping (a classification task), under particularly challenging conditions. These include settings with extremely limited sample sizes, unpaired training data across modalities, and cases where certain modalities are missing for specific classes, as is often the case with rare diseases. Most of the approaches for multi-omics data integration \cite{argelaguet2021computational} require large datasets for their training and are thus not directly applicable to this difficult setting. To extend multi-omics data integration applicability to rare diseases or other types of diseases with scarce data, we propose to take advantage of large public datasets of multi-omics data such as the TCGA database \cite{weinstein2013cancer} to help with the integration. 

From a machine learning perspective, we consider that the training dataset includes data from multiple modalities (the various -omics) and for several classes (the various diseases, cancer, conditions or subtypes) from the available databases. Yet it is often the case that the training dataset is not complete in multiple ways. First, the samples are only observed in some of the modalities, i.e. they are unpaired. Moreover a particular modality/class pair may be missing in the training set, meaning that there are no examples of this modality in the training samples for a given condition (the class). Finally a number of samples may lack a class label, since labeling requires expensive expert annotation. The goal of this study is to propose solutions for multi-omics data integration within this context.

To handle this situation effectively, the model must be designed with specific features. First, the model should be able to generate a representation of the data samples that is independent of the modalities from which a given sample come from. In addition, within this representation, it should be possible to discriminate the class to which a sample belongs. This requires being able to align the samples observed in one modality to samples observed in another modality, even though the training data samples are unpaired across modalities. Second, the method should be able to exploit samples with no class labels information. Third, the classifier should be learnable for a set of classes with potentially just a few training samples, hence it should be able to deal with an imbalanced classification problem \cite{ochal2023few}. To answer these three challenges, we propose an approach named MODIS for Multi-Omics Data Integration for Small and unpaired datasets. 

\begin{figure}[!htp]
    \centering
    \includegraphics[width=0.7\textwidth]{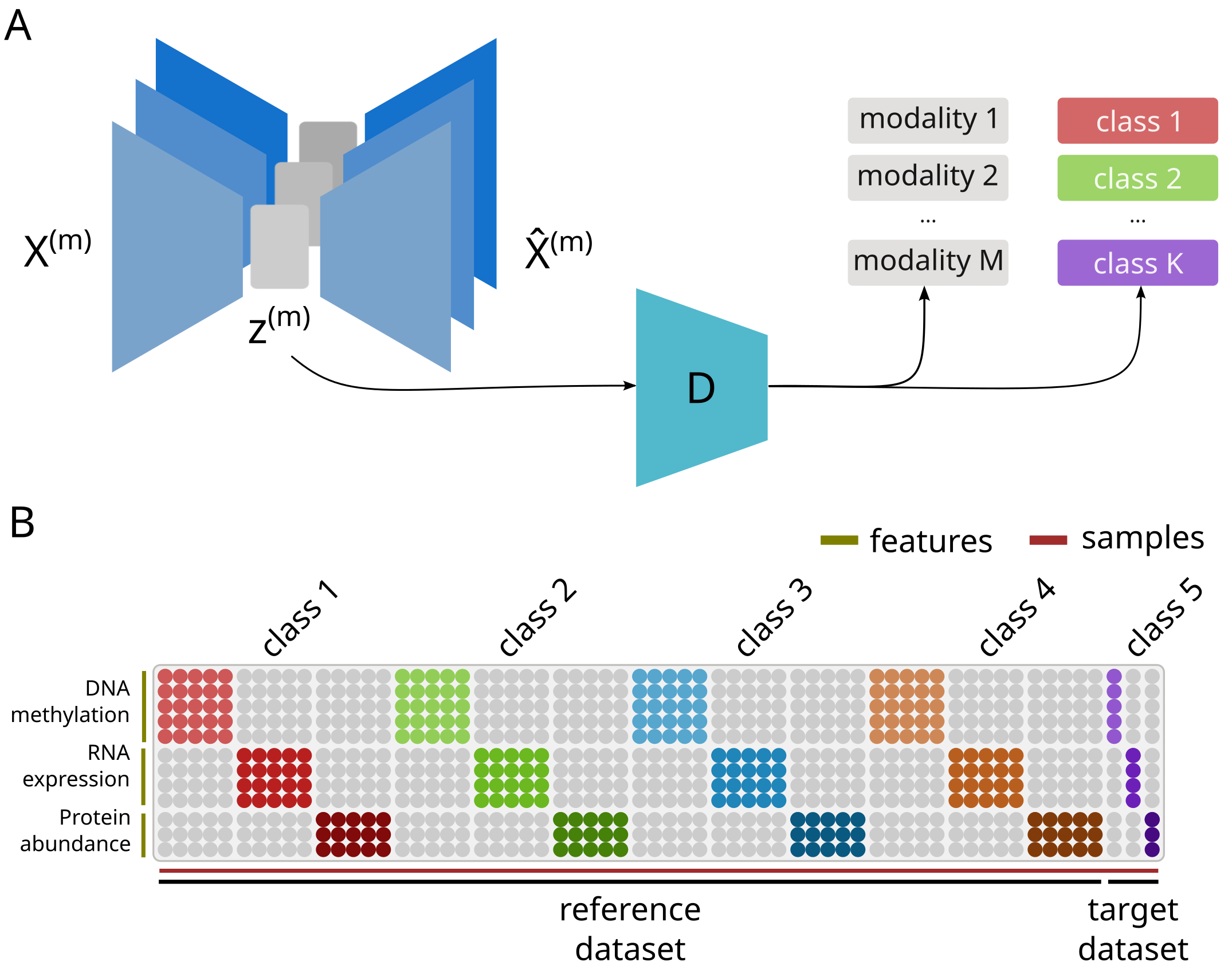}
    \caption{\textbf{Schematic of the MODIS architecture and our proposed approach to address the challenge of learning with small datasets as a class imbalance problem}. \textbf{A}, Each modality is encoded by a specific variational autoencoder (VAE). The embeddings generated by each VAE are integrated and aligned in a shared latent space using an adversarial learning approach based on an auxiliary modality discriminator D. The auxiliary discriminator D is additionally used for class learning, trained with full or semi-supervision. The implementation is described in details on Fig.~\ref{supp_model}. \textbf{B}, A large reference dataset is combined with a small target dataset into an imbalanced unpaired multi-omics dataset leveraging the knowledge from the reference to improve the alignment of the target dataset. Each column within a modality represents a sample, and each row corresponds to a feature of that modality. Gray dots denote unobserved samples.}
    \label{Fig1}
\end{figure}

Several approaches have been dedicated to multiview or multimodal data with the goal of performing a prediction task, e.g., classification, from a number of very different and complementary modalities whose heterogeneity prevents any simple combination (see for example \cite{DBLP:conf/iclr/HanZFZ21}). More generic approaches have been proposed that combine modalities, such as audio and text, in the CLAP model \cite{DBLP:conf/icassp/ElizaldeDIW23}, to learn rich representations that may be used for (any) downstream task. Another line of works have focused on extending data analysis classics such as canonical correlation analysis (CCA) to explore relationship between different modalities \cite{DBLP:conf/icml/AndrewABL13}. Besides, a number of works have been proposed to learn to infer a modality from another one, usually in a supervised setting, in a general machine learning setting \cite{NgiamKKNLN11} and for multi-omics data \cite{yang2021multi}. A common backbone of these methods consists in a set of auto-encoders, one for each modality, that are learned in such a way that their latent representations are aligned, thus enabling prediction of a modality from another one \cite{NgiamKKNLN11,yang2021multi} by encoding with a modality encoder and decoding with another modality decoder. Learning to align the latent representation spaces of multiple modalities autoencoders can be done using paired data (all modalities are observed for each training samples) if this data is available \cite{tang2023explainable}. Alternatively, in the case of unpaired data where only one or some of the modalities are available for the training samples, alignment can be obtained through distribution constraints \cite{yang2021multi,he2024mosaic}, extending the seminal work from \cite{DBLP:journals/corr/MakhzaniSJG15}. These approaches are based on generative adversarial networks, which are notoriously difficult to tune but have recently benefited from a novel loss function that improves their stability during training \cite{huang2024gan}.

In computational biology and bioinformatics, the question of integrating multiple modalities for classification or clustering has been addressed by various authors in the past few years. In the case of paired samples across modalities, several linear methods such as joint matrix factorization have been used, as reviewed in \cite{cantini2021benchmarking}, and extended to small dataset settings with transfer learning \cite{hirst2025motl}. Recently, non-linear methods such as multiple auto-encoders with a shared latent space have been proposed, for paired samples (also known as vertical integration \cite{argelaguet2021computational}) \cite{tang2023explainable}, unpaired samples (diagonal integration) \cite{yang2021multi,samaran2024scconfluence}, or when considering a combination of paired or unpaired samples (mosaic integration) \cite{he2024mosaic}.

Few approaches address imbalanced classification problems \cite{huang2016learning} or handle learning classes with very few training samples (few shot learning \cite{wang2020generalizing}). The question of combining multi-omics data integration with the question of class imbalance remains largely unexplored.

\section{Methods}

As stated in the introduction MODIS aims to perform multi-omics data integration and classification with small and unpaired datasets. We propose to train coupled autoencoders (Fig. \ref{Fig1}A) on both a large reference dataset and a small target dataset (Fig. \ref{Fig1}B), the latter corresponding for instance to a rare disease dataset. This approach aims at leveraging the structure learned from the larger dataset to improve the alignment, translation of modalities, and label classification in the smaller dataset. 

MODIS is composed of multiple coupled variational auto-encoders (VAEs), as many as the number of modalities. To link the samples from different modalities, we align the latent representation of each of the modality encoders. To do so, we enforce the latent representation of each sample, computed by the corresponding modality encoder, to be \textit{modality-free}, following \cite{DBLP:journals/corr/MakhzaniSJG15,yang2021multi,huang2024gan}. In practice, we use an auxiliary adversarial modality discriminator that operates in the latent space. The auxiliary adversarial modality discriminator is a classifier trained to recognize the modality of a sample from its latent representation, while the encoders of each of the modalities are trained to fool this classifier \cite{goodfellow2014generative, mao2017least,lim2017geometric}. Doing so, one aligns the latent representation of all modality autoencoders so that the latent space becomes a shared latent space. This is a key aspect of our method since it enables both the possibility of reconstructing a modality from another one, by chaining the encoder of a modality with the decoder of another modality, and the possibility of learning and using a single classifier for the classes independently of the observed modality, since it operates in the shared latent space. We are particularly interested in datasets with multiple biological conditions which are modeled as multiple classes in the data. To ensure class separability, we incorporated an auxiliary class discriminator operating on the latent space and during training, adapted from the AC-GAN architecture \cite{odena2017conditional}. Moreover, we added a regularization term that favors clustered latent representations, adapted from \cite{tang2023explainable,kampffmeyer2019deep}. Overall the architecture of MODIS aims at preserving the underlying data structure while ensuring meaningful alignment across modalities and class separability, see Fig.~\ref{Fig1} and Fig.~\ref{supp_model}.

Since the training involves an adversarial discriminator, the optimization iteratively alternates two steps. In the first step, only the  discriminator parameters are updated, to improve modality and class identification from the latent space using the following loss 
\begin{equation}
   L_D = L_{\text{relativistic},D} + L_C + L_{\text{clustering}}\label{eqn_D}
\end{equation}
where  $L_{\text{relativistic},D}$ is adapted from the regularized relativistic loss for GAN (R3GAN) \cite{jolicoeur2018relativistic, huang2024gan} and aims at training the modality classifier $D_{\text{mod}}$ to predict modality from the latent representation, $L_C$ is the cross-entropy loss which aims at training the auxiliary class classifier $D_{\text{class}}$ for class prediction from the latent representation, 
and, following \cite{tang2023explainable, kampffmeyer2019deep}, we used a clustering loss, noted $L_{\text{clustering}}$, that ensures the separability and the compactness of the class clusters in the latent space. The loss functions $L_{\text{relativistic},D}$, $L_C$ and $L_{\text{clustering}}$ are described in details in the supplementary material (\ref{modis_training}).

In the second step, the discriminator parameters are frozen and all other parameters (all the VAEs parameters) are updated using a combined loss. The combined loss includes all VAEs standalone losses (reconstruction losses $L_{\text{recon}}^{(i)}$ and Kullback-Leibler divergences $D_{KL}^{(i)}$ for each modality $i$), a classification (adversarial) loss ($L_{\text{relativistic},\text{VAE}}$), and the label classification and clustering losses ($L_C + L_{\text{clustering}}$) as above.  

\begin{equation}
   L_{\text{VAE}} = \sum_{i = 1}^{M}\left(L_{\text{recon}}^{(i)}+ \beta D_{KL}^{(i)}\right)+ L_{\text{relativistic},\text{VAE}} + L_C + L_{\text{clustering}}  \label{eqn_VAE}
\end{equation}
where $M$ is the number of modalities, $\beta$ is an hyperparameter, $L_{\text{relativistic},\text{VAE}}$ is the \textit{adversarial} regularized relativistic loss of the discriminator with respect to the modalities (a variant of $L_{relativistic,D}$ for training the encoders such that they fool the modality classifier $D_{\text{mod}}$, as is standard in adversarial learning \cite{DBLP:journals/corr/MakhzaniSJG15,huang2024gan}). The various loss functions $L_{\text{recon}}$, $D_{KL}$, $L_{relativistic,VAE}$, $L_C$, $L_{\text{clustering}}$ are detailed in the supplementary material (\ref{modis_training}). The dimensions of the neural networks used in practice in the experiments are reported on Fig.\ref{supp_model}.

This approach extends naturally to a semi-supervised setting where the model is trained on a combination of data samples whose class label is known and data samples whose class label is unknown. The relative proportion of labeled and unlabeled data samples required is analyzed in the results section.

\subsection{Datasets}

\paragraph{InterSIM Dataset} To test our approach, we generated realistic multi-omics datasets by simulation using the \textbf{InterSIM} $R$ package \cite{chalise2016intersim}. The three -omics represented are DNA methylation (367 features), RNA expression (131 features), protein abundance (160 features). We started by creating a collection of large paired multi-omics datasets with 11,500 samples (each sample being a triplet of -omics) and distributed among 5 balanced classes (each class representing $\sim$ 20 \% of the samples). The datasets within this collection differ by their cluster mean shift values ($d$), this parameter encodes the separability of the clusters \cite{chalise2016intersim}. We chose for cluster mean shift values $d$ equal to 0.5, 1, 1.5, 2, 3. The corresponding datasets are represented on Fig. \ref{intersim_datasets}. To simplify the presentation, we chose to highlight our key results on the dataset corresponding to $d = 2$ in the main text of this article. From the 11,500 multi-omics paired samples (given a value of cluster mean shift), we derived an unpaired dataset by keeping only one representative -omic modality per sample while preserving the balanced distribution among the 5 classes. We finally used the \verb|train_test_split| function of \verb|sklearn| with a ratio of 0.2 and balanced classes to create the train/test datasets, and performed a standard scaling transformation of the train set and applied the learned transformation to the test set to prevent data leakage. We adapted these datasets to the various experimental settings (semi-supervised, class imbalance, missing modality). To evaluate the class prediction performance of MODIS, we used the balanced accuracy score (B-ACC), which is defined as the average of recall obtained on each class, the normalized mutual information (NMI), and the Adjusted Rand Index (ARI) \cite{pedregosa2011scikit}. To evaluate the reconstruction and translation between modalities, we used the mean squared error (MSE). When considering the task of translation, which corresponds to the composition of an encoder for one modality and a decoder for another modality, since we originally generated paired samples, given a sample in one -omic, we have the ground truth in each of the other -omics. The translation error is computed between the predicted -omic and the ground truth.

\paragraph{TCGA Dataset} We used Python scripts interfacing with the Genomic Data Commons (GDC) Data Portal API (release 45.0) to download DNA methylation, mRNA expression, and miRNA expression data for each patient in the TCGA cohort. DNA methylation data consist of CpG site $\beta$-values, while mRNA and miRNA expression data are raw counts.

The data was then pre-processed using Python. For DNA methylation data, we only used samples sequenced with the Illumina Human Methylation 450 platform. We removed probes that map to sex chromosomes or that do not map to CpG sites. We also removed probes with missing values or zero expression in $\geq 20\%$ of the samples, as well as those with a standard deviation $< 0.1$. We imputed missing values with the mean value of the probe. Finally, we transformed the $\beta$-values to M-values. For both mRNA and miRNA expression data, we removed genes if they had less than 5 counts or missing values in $\geq 90\%$ of the samples. We normalized the counts using a custom Python implementation of the DEseq2 median of ratios method and subsequently log2(x+1) transformed the normalized counts \cite{hirst2025motl}.

Due to the mosaic nature of the data, a custom Python script was used to split the samples into train and test with a ratio of 0.2 and stratified by class for each modality combination (Table~\ref{tab:tcga_data}). Three overlapping subsets of the training and test sets were generated, containing 10, 20, and 34 classes, respectively. Each subset is composed of multiple classes, each corresponding to samples of a given cancer type, and a normal tissue class, composed of normal tissue samples. Classes were assigned in order of decreasing sample size, see Fig.~\ref{Fig5}.

To reduce the dimensions of the data, we standardized and normalized the train dataset and applied PCA to the three omics using \textit{scikit-learn} library in Python. We retained the top 2,500 components, capturing 88.9\% and 88.7\% of the variance, for DNA methylation and mRNA expression respectively. For miRNA we used the entire set of 542 features. After dimensionality reduction, we standardized and normalized again the train dataset, which then serves as an input for MODIS training. We used the principal components obtained in the train dataset to reduce the dimensionality of the test dataset in order to avoid data leakage, followed by a standardization and normalization.

To determine the optimal training parameters (listed on Table~\ref{tab:model_hyperparameters}), we performed a full grid search with 5-fold cross-validation on the 10-class subset of TCGA. The hyperparameters tuned were the beta value, latent size, learning rate, and the lambda parameter for the zero-centered gradient penalties in the relativistic loss. Model performance was evaluated considering the balanced accuracy (B-ACC) of the discriminator predictions and the reconstruction loss of the VAEs, measured as the MSE.

MODIS was trained over 4000 epochs and we used the parameters of the epoch obtaining the best global loss. An example of training dynamics in one of the most challenging situation is shown on Fig. \ref{training_curves}, for the dataset TCGA 34 classes. 

\subsection{Comparison to another multi-omics integration method}
To evaluate the performances of MODIS against a state-of-the-art multi-omics data integration method \cite{liu2025multitask}, we chose to run MIDAS (v0.1.15 - \cite{he2024mosaic}) on the same TCGA datasets. MIDAS is a deep probabilistic model for mosaic and diagonal data integration. MIDAS, like MODIS, has the advantage of being able to integrate three modalities and provide a joint latent embedding. We implemented wrapper functions to allow MIDAS to run with the same data loaders used for MODIS in order to provide a comparable setting. The data was not standardized a second time as with MODIS since it resulted in a severe performance drop. We used the same training dataset as for MODIS. The configuration parameters used are: \texttt{dim\_c = 512}, \texttt{dims\_shared\_enc = [3000, 1500]}, \texttt{dims\_shared\_dec = [1500, 3000]}. In addition, in MIDAS parameters, following the setting in \cite{he2024mosaic}, Poisson sampling was used for mRNA and miRNA modalities while Gaussian sampling was used for DNA methylation modality. The modality specific \texttt{lam\_recon} parameter was set to 1 in all cases. The models were trained for up to 3,000 epochs. To evaluate classification accuracy, the k-means algorithm was applied on the latent representation of the test datasets. To assign the resulting classes to the ground truth we used the hungarian algorithm \cite{kuhn1955hungarian}. Balanced accuracy (B-ACC), normalized mutual information (NMI) and adjusted rand index (ARI) metrics were computed on the results. MIDAS was evaluated on the 10, 20 and 34 classes subsets of TCGA.


\section{Results}

\paragraph{Latent space alignment}

\begin{figure}[!htp]
    \centering
    \includegraphics[width=0.90\textwidth]{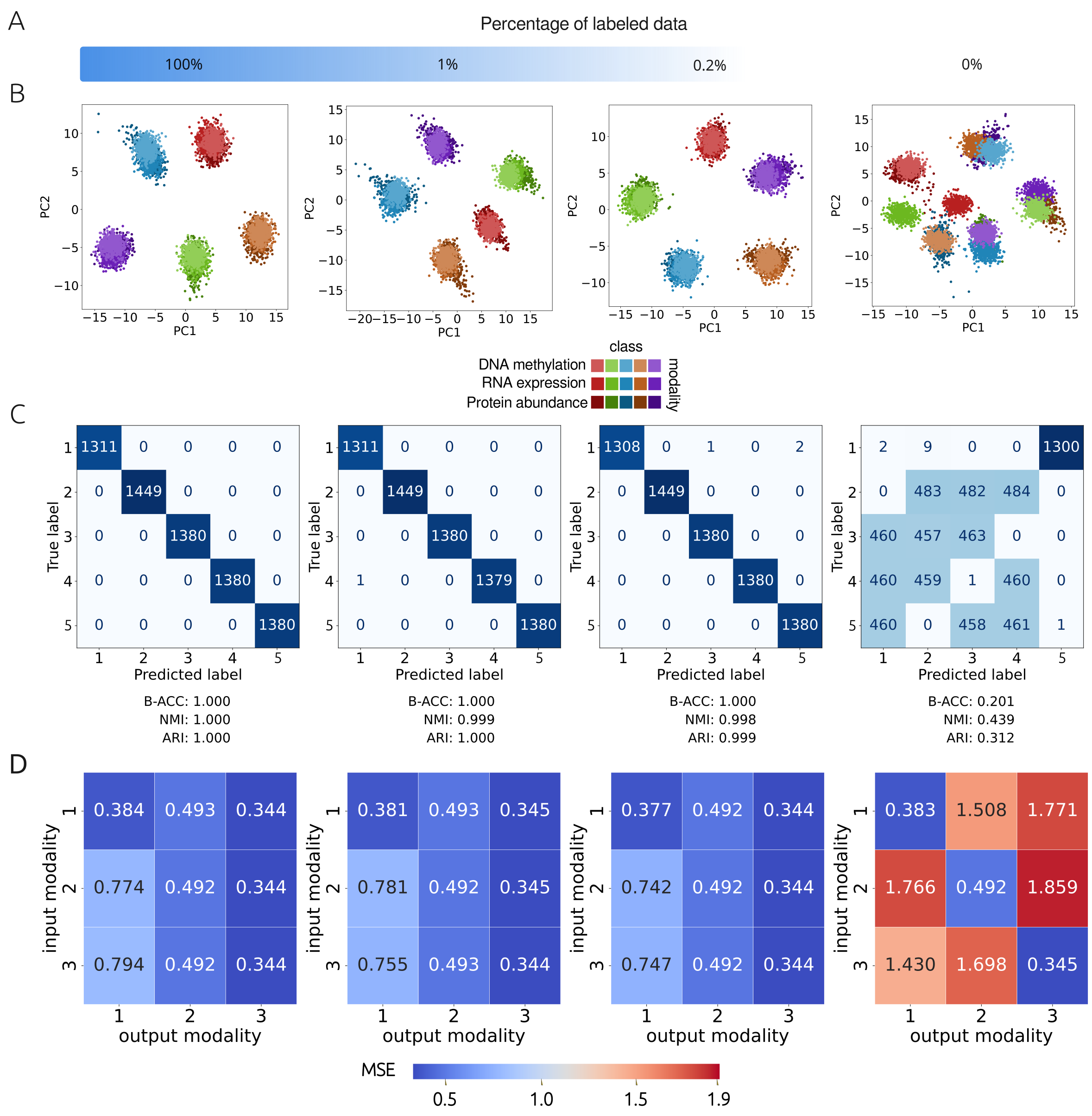}
    \caption{\textbf{Latent space alignment as a function of available class labels}. MODIS has been trained under four levels of supervision using the InterSIM dataset, each column correspond to a level of supervision which is associated to a given amount of labeled data. \textbf{A}, Percentages of class labels used in each of the four levels of supervision (each column). \textbf{B}, PCA projection of the test set latent embeddings \textbf{C}, Confusion matrices of the discriminator's class predictions. \textbf{D}, Heatmaps with the MSE for the modality reconstruction and translation between MODIS prediction and ground truth on the test set (Modality 1: DNA Methylation, Modality 2: RNA expression, Modality 3: Protein abundance).}
    \label{Fig2}
\end{figure}

MODIS consists of coupled auto-encoders sharing a common latent representation constructed such that the modalities of origin of the samples are indistinguishable while the classes from which each sample belongs to are easily identifiable.

To establish a baseline for the quality of the latent space alignment between modalities, we first trained our architecture on a fully supervised (i.e., class labels available for all samples) and class-balanced dataset (Fig. \ref{Fig2}, first column). In this ideal setting, the model achieved 100\% class prediction accuracy on test data using the trained class discriminator, demonstrating its capacity to learn a well-structured latent representation when provided with full supervision. Moreover, the reconstruction error and translation error computed as the MSE were down to 0.34-0.79 (Fig. \ref{Fig2}.D), confirming that 1) the latent representation of each modality is sufficient for the faithful reconstruction of the data and 2) the latent representations between modalities are sufficiently well-aligned so that the latent encoding of one modality leads to faithful reconstruction in another modality. Of course, one cannot achieve any meaningful alignment between modalities in the absence of class labels, i.e. with no assignment constraint between the clusters of the various modalities, which is shown in the right-most column of Fig. \ref{Fig2}. In this case, the model did recover the clustered structure of the data, but as expected, struggled to match the clusters between the various modalities, leading to random class prediction (class prediction accuracy of 20\% across five classes). And similarly, as expected, the reconstruction error is as low as in the fully supervised case, with value around 0.4 for each of the three modality, but the translation error is much higher, e.g. with value 1.85 from modality 2 to modality 3. This confirms the misalignment of the modalities in the latent space. Overall, these results highlight the importance of explicit guidance in ensuring a well-aligned latent space across modalities.

We explored an intermediate, semi-supervised, setting in which only a subset of the training samples were annotated with their class labels. Our results show that even with minimal supervision, where only 1\% or as little as 0.2\% of the training data were annotated with their class label, our architecture was able to recover a nearly perfect class prediction accuracy on test data (Fig. \ref{Fig2}.C, second and third column) and a reconstruction and translation error almost identical to the fully supervised baseline (Fig. \ref{Fig2}.D, second and third column). Interestingly, the proportion of 0.2\% of labeled data correspond to the extreme case where only one labeled sample is available per class/modality pair. We evaluated the behavior of this semi-supervised setting with various degrees of variability in the simulated datasets, Figs. \ref{intersim_2.0d},\ref{intersim_1.5d},\ref{intersim_1.0d}, and \ref{intersim_0.5d}. We note that for a cluster mean-shift value (d) equal to 0.5, the classification accuracy drops when only 0.2\% is annotated. However, the very good classification accuracy is recovered when the supervision is increased to 1\% of the training data. These results shows that MODIS is stable even in the presence of variability in the training data, while however requiring a slight increase in the amount of supervision to align the latent space and provide a good reconstruction and translation error in more challenging contexts.

\paragraph{Handling small datasets with class imbalance}

\begin{figure}[!htp]
    \centering
    \includegraphics[width=0.95\textwidth]{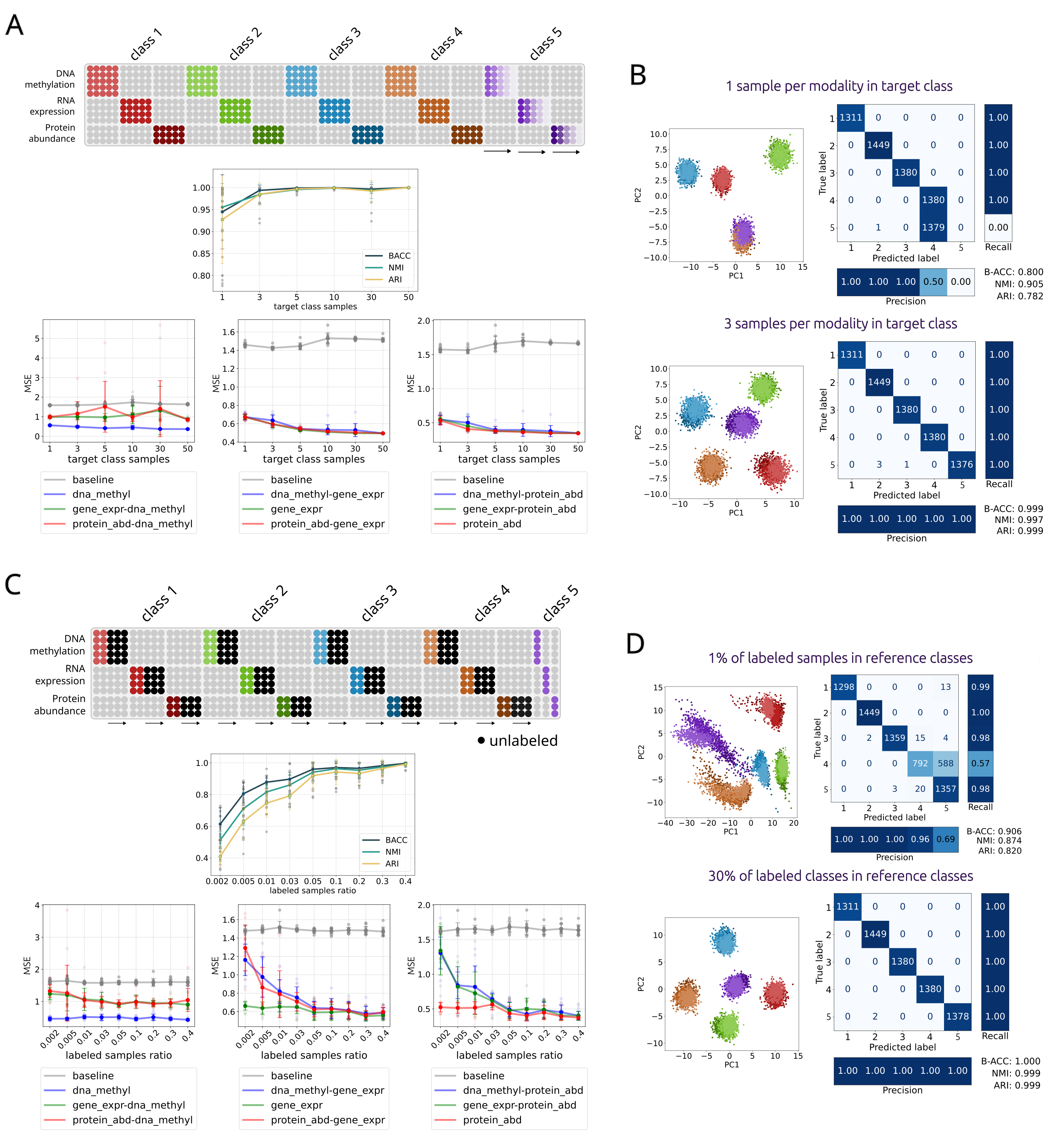}
    \caption{\textbf{Class imbalance between a reference dataset and a target dataset in fully supervised and semi-supervised settings.} Arrows indicate the groups of samples whose size is varied during evaluation. \textbf{A–B}, Class imbalance with full class label supervision. \textbf{A}, Evaluated setting, followed by class prediction performance (balanced accuracy (BACC), normalized mutual information (NMI), and adjusted rand index (ARI)) as a function of the number of training samples in each of the modality in the target class (class 5). Also shown is the MSE (mean ± standard deviation) for modality reconstruction and translation between MODIS predictions and ground truth, as a function of the number of target class samples. The baseline corresponds to the MSE between all sample pairs within each modality. Each data point reflects the mean over 10 replicates. \textbf{B}, PCA projection of the test set latent embeddings and confusion matrices of MODIS class predictions with 1 and 3 training samples per modality in the target class. \textbf{C–D}, Class imbalance under partial class label supervision (semi-supervised setting). The target class training set is fixed to 18 samples (6 per modality). \textbf{C}, Evaluated setting, followed by the same plots as in panel A, but evaluated as a function of the proportion of labeled samples in the reference training dataset. Each data point again reflects the mean over 10 replicates. \textbf{D}, PCA projection of the test set latent embeddings and confusion matrices of MODIS class predictions trained with 1\% and 30\% labeled samples in the reference dataset.}
    
    \label{Fig3}
\end{figure}

In biomedical contexts, training datasets are often small, particularly in the case of rare diseases, making it challenging to develop robust predictive models \cite{banerjee2023machine}. As is common in machine learning, it can be beneficial to exploit shared information from a larger reference dataset and combine it to improve the performance on the smaller target dataset \cite{hirst2025motl}. With MODIS, we propose to address this aspect by training our model on both a large reference dataset containing multiple classes of unpaired samples and a small target dataset containing one class of unpaired samples Fig. \ref{Fig1}.B. 

To evaluate the effect of class imbalance on MODIS predictive performance, we first considered a setting in which the reference dataset contained a large number of samples across multiple classes, while the target dataset consisted of only a few samples in a specific class (Fig. \ref{Fig3}A). We evaluated the impact of sample size on classification performance by analyzing the evolution of class prediction accuracy as a function of the number of training samples for the target dataset. Our results indicate that, even with a low number of samples, the model achieves excellent class prediction accuracy. Specifically, with 5 training samples per modality in the target dataset (0.2\% of the whole training dataset), MODIS achieves near perfect class prediction accuracy (Fig. \ref{Fig3}A). With 1 training sample per modality in the target dataset, MODIS achieves 95\% balanced class prediction accuracy. When looking at the class prediction, (Fig. \ref{Fig3}B), one can see that having only one training sample per modality can lead to deteriorated class prediction. Three training samples per modality however, can already lead to near perfect accuracy at the single class level. Naturally, the class prediction accuracy grows with the increase in the number of samples in the target dataset. This suggests that the knowledge transfer from the large reference dataset plays a crucial role in stabilizing the representation of the target class, allowing for accurate predictions despite the extremely small dataset size.  

Next, we investigated the minimal level of class supervision required in this highly imbalanced setting (Fig. \ref{Fig3}C). In this case, we fixed the number of samples in the target dataset to 6 per modality (18 in total) and systematically varied the proportion of class-labeled versus unlabeled samples in the reference dataset. Our results reveal that the percentage of annotated data in the reference dataset influences the class prediction accuracy of both the reference and the target dataset. Notably, we found that the class prediction accuracy could exceed 0.95 with only 5\% of the reference dataset samples being annotated, and increases with the proportion of labeled samples. This finding underscores the efficiency of our semi-supervised approach, demonstrating that even a small amount of supervision in the reference dataset is sufficient to enable accurate classification, even  in an imbalanced setting. This suggests that, in an applied scenario, it could be interesting to increase the size of the reference dataset with unlabeled data, which can help improve prediction. Overall, these results highlight the potential of leveraging large, not necessarily well-annotated datasets to enhance model performance, a crucial advantage for rare disease research where data is often scarce.  

\paragraph{Assessing the impact of missing modality for a given class}

\begin{figure}
    \centering
    \includegraphics[width=\textwidth]{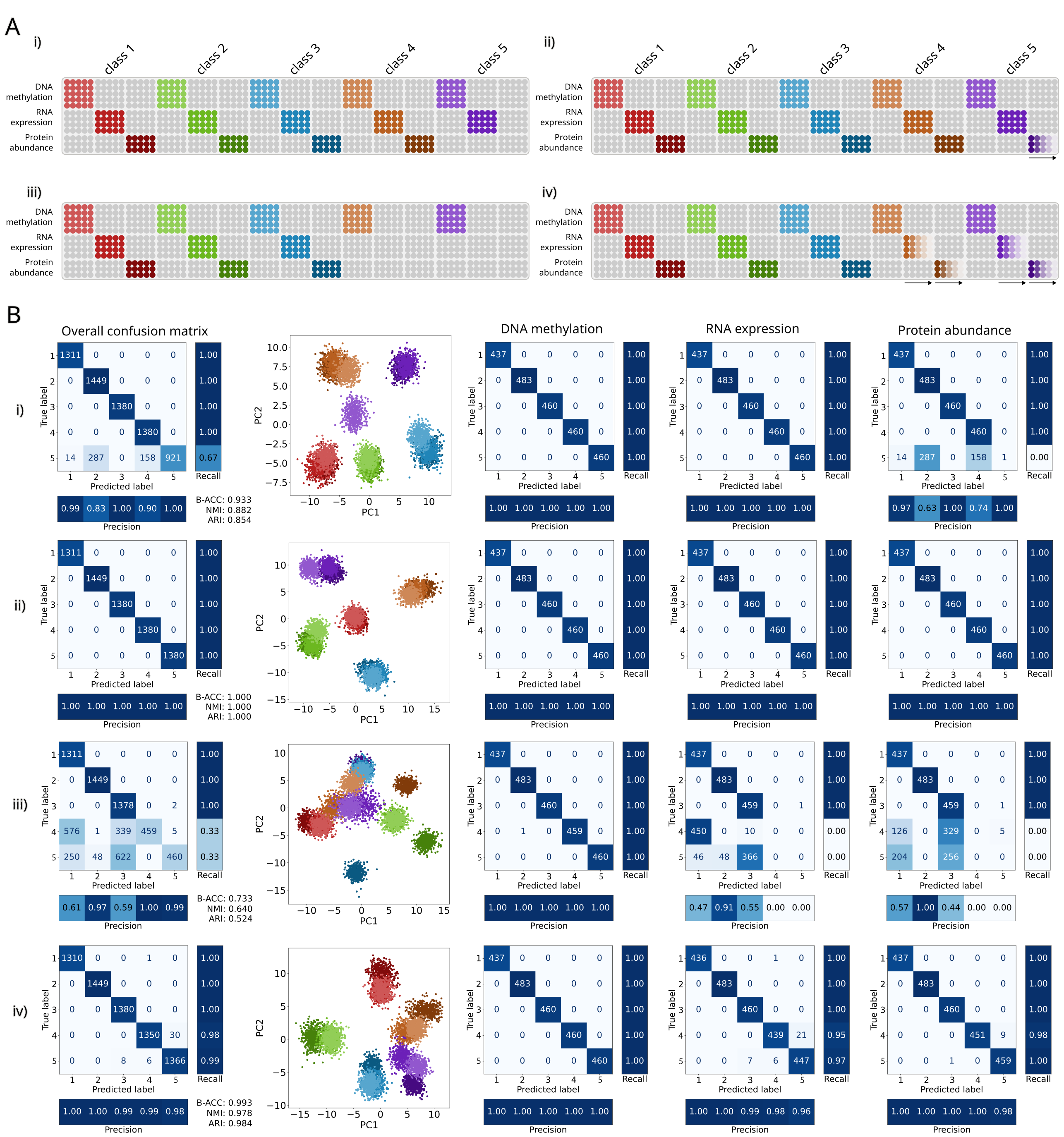}
    \caption{\textbf{Effect of missing or partially missing modalities on latent space alignment and class prediction performance.} \textbf{A}, Evaluated settings for the training set: \textbf{i)} one class/modality pair is missing; \textbf{ii)} the number of training samples is varied in the class/modality pair missing in \textbf{A-i}; \textbf{iii)} four class/modality pairs are missing across two classes; \textbf{iv)} the number of training samples is varied in the four class/modality pairs missing in \textbf{A-iii} \textbf{B}, MODIS results on each of the settings presented in \textbf{A-(i-iv)}. Columns: (1) overall confusion matrix of MODIS class predictions, (2) PCA projections of test set embeddings in the joint latent space, (3–5) modality-specific confusion matrices for DNA methylation, RNA expression, and protein abundance, respectively. In \textbf{A-ii} and \textbf{A-iv}, the evaluated class–modality pairs contain only a single sample. Complementary classification prediction performance, reconstruction error, and translation error as a function of the number of training samples in these small class–modality pairs are shown in Fig.~\ref{supp_eval_intersim}.}
    \label{Fig4}
\end{figure}

Beyond the issue of class imbalance, we also investigated the impact of missing modalities in one or more classes on the performance of our model. To systematically evaluate this challenge, we designed a series of increasingly difficult settings and assessed the model ability to maintain latent space alignment and accurate class prediction despite missing data (Fig.~\ref{Fig4}.A).

In the first case, we removed all samples from one modality in a single class to observe whether the model could still accurately classify samples from the missing modality (Fig.~\ref{Fig4}.A-i). Despite the complete absence of data for that modality-class pair, the model achieved a balanced class prediction accuracy of 93\% (Fig.~\ref{Fig4}.B-i). When looking closely at the class level prediction performance, one can see however that in this setting, the model fails in prediction mode for the class/modality pair absent from the training dataset. Beyond this class/modality pair, the model performs with nearly perfect accuracy. While MODIS successfully leverages information from the available modalities and classes to infer meaningful representations, allowing for accurate predictions even in the presence of missing data, it does not recover a reliable representation of the missing class/modality. To further explore the impact of introducing minimal data in the missing modality, we gradually increased the number of samples available in that modality-class pair (Fig.~\ref{Fig4}.A-ii). We found that adding just a single sample was sufficient to restore the perfect class prediction accuracy (Fig.~\ref{Fig4}.B-ii, Fig.~\ref{supp_eval_intersim}A), indicating that even a minimal number of samples can significantly improve model performance in such cases. Adding more training samples for the considered class/modality pair helps with reconstruction and translation accuracy.

Next, we tested a more challenging scenario in which two modalities were missing in two different classes (Fig.~\ref{Fig4}.A-iii). As expected, this setting led to a reduction in overall model accuracy, with a global class prediction balanced accuracy of 73\% (Fig.~\ref{Fig4}.B-iii). Notably, while the model still performed well across the dataset, the accuracy dropped for the four class/modality pairs missing from the training dataset. To determine whether the introduction of even a small number of samples could mitigate this effect, we incrementally added samples to the missing modalities  (Fig.~\ref{Fig4}.A-iv, Fig.~\ref{supp_eval_intersim}B). Once again, we observed a substantial improvement, with the overall accuracy increasing to 96\% after the addition of just one sample per missing modality-class pair. These findings emphasize the robustness of our approach in handling incomplete multi-modal datasets while also highlighting the critical role of even minimal training samples in restoring prediction accuracy. This capability is particularly relevant in real-world biomedical applications, where data collection across multiple modalities is often incomplete or imbalanced across different classes.

\paragraph{TCGA}

Our experiments thus far have been conducted using simulated data \cite{chalise2016intersim}, which provides a controlled environment for evaluating model performance. However, an important question remains: how well does our approach generalize to real-world data? To assess its applicability in a more complex and biologically relevant setting, we applied our architecture to The Cancer Genome Atlas (TCGA) dataset. This large-scale database contains multi-omics data from various cancer types, making it an ideal benchmark for evaluating cross-modal integration with imbalanced classes in a real-world scenario.

\begin{figure}[!htp]
    \centering
    \includegraphics[width=\textwidth]{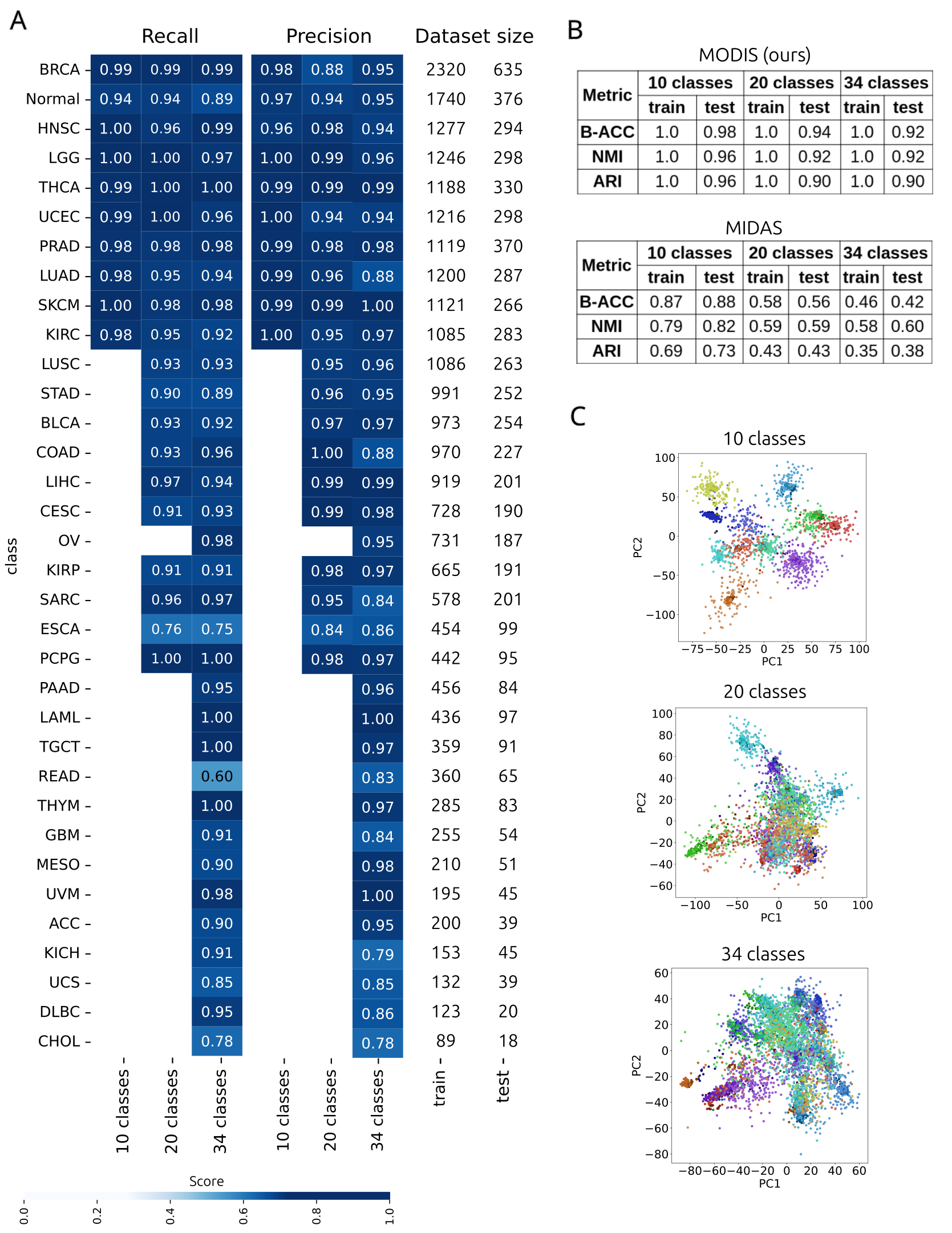}
    \caption{\textbf{Evaluation of MODIS on TCGA test subsets of 10, 20, and 34 classes, consisting of distinct cancer types and normal tissue.} \textbf{A}, Heatmap of recall and precision per class for each subset, along with the number of samples in the train and test datasets. \textbf{B}, Class prediction performance (balanced accuracy (BACC), normalized mutual information (NMI), and adjusted rand index (ARI)) for train and test datasets. \textbf{C}, PCA projection of the test set latent embeddings for each subset on the test dataset.}
    \label{Fig5}
\end{figure}

This diverse dataset introduces additional challenges compared to simulated data, including higher variability, technical noise, and potential batch effects across modalities. To assess the performance of MODIS under these conditions, we tested it on different TCGA subsets containing 10, 20, and 34 classes, where the normal tissue class was consistently included and the remaining classes were selected in decreasing number of sample (Fig.~\ref{Fig5}). The 34-class set corresponds to the complete TCGA dataset. This multimodal dataset encompasses three omics modalities: DNA methylation, RNA expression, and miRNA expression (Table~\ref{tab:tcga_data}).

Despite the dataset complexities, MODIS achieved 100\% accuracy on the training sets across all subsets, underscoring its ability to learn highly discriminative latent representations (Fig.~\ref{Fig5}B). More importantly, when evaluated on unseen test data, the model maintained strong global prediction accuracies of 98\%, 94\%, and 92\% for the 10-, 20-, and 34-class subsets, respectively, demonstrating robust generalization beyond the training data. The decrease in accuracy across subsets can be attributed to the increasing number of classes with fewer samples. As expected, recall and precision were lower for these underpopulated classes, but their behavior differed: recall declined as the subsets grew larger, whereas precision was more affected by misclassification in the smaller classes (Fig.~\ref{Fig5}A). Overall these results confirm that MODIS effectively integrates heterogeneous modalities to classify biological conditions with high accuracy on real experimental datasets (Fig.~\ref{Fig5}C). In comparison, state-of-the-art MIDAS \cite{he2024mosaic}, trained on the same dataset, led to lower performances. Specifically, the model led to global prediction accuracies of 88\%, 56\%, and 42\% for the 10-, 20-, and 34-class subsets, respectively. These lower performances Interestingly, MODIS obtained low reconstruction error, despite the increasing number of classes, Fig. \ref{MODIS_reconstruction_error}.

\begin{figure}
    \centering
    \includegraphics[width=\textwidth]{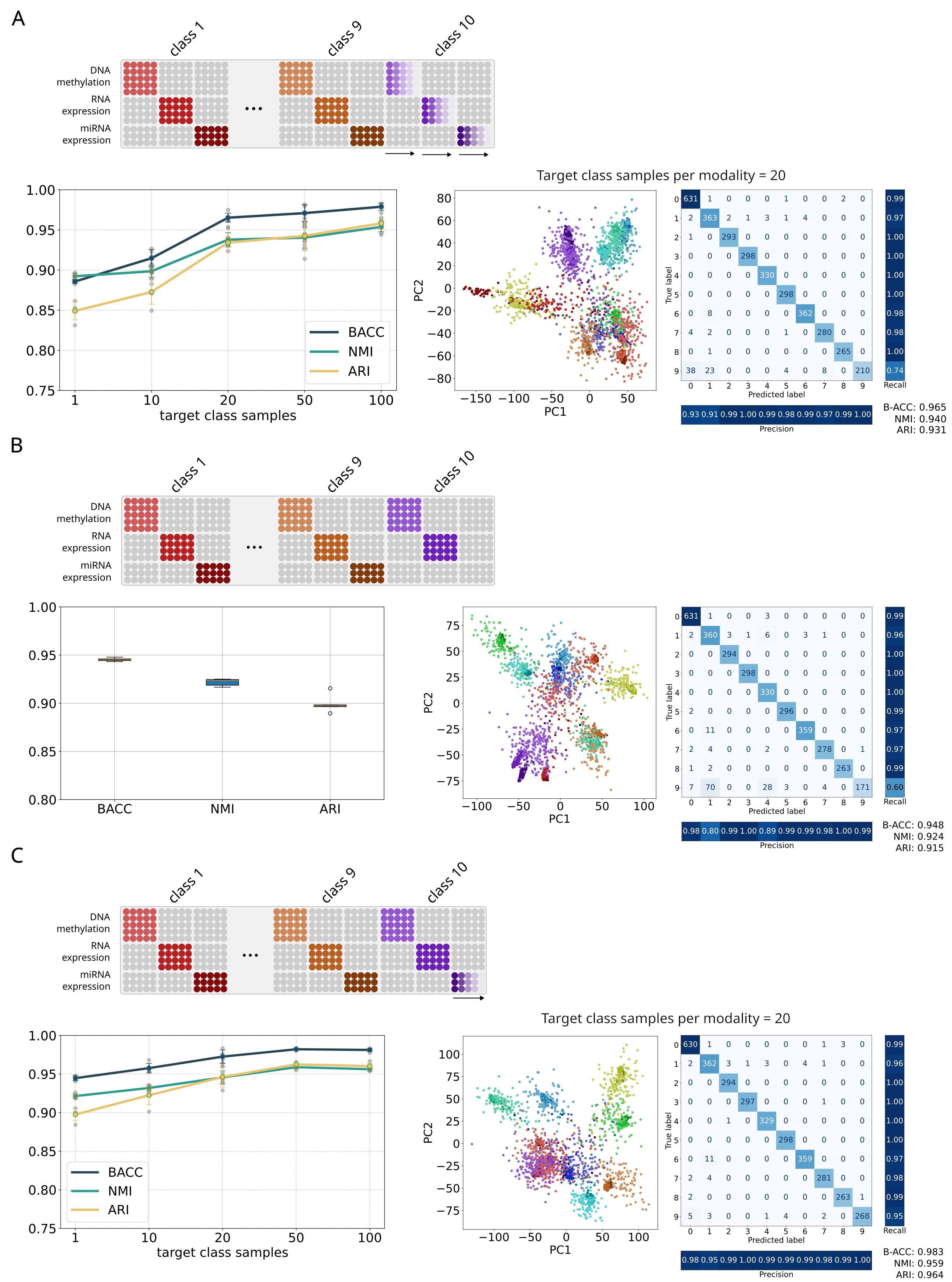}
    \caption{\textbf{Evaluation of MODIS on the test TCGA dataset across 10 classes (nine cancer types and one normal tissue class) under varying conditions.}
    \textbf{A}, Performance under class imbalance, \textbf{B}-\textbf{C}, Performance under missing and partially missing modality settings, respectively. The top row of each panel shows the evaluated setting. The bottom row of each panel contains three columns: (i) a scatterplot or boxplot of the results showing the class prediction performance (balanced accuracy (BACC), normalized mutual information (NMI), and adjusted rand index (ARI)), all data points have 5 replicates, (ii) PCA projection of the test set latent embeddings, and (iii) the discriminator’s confusion matrix. Panels \textbf{A} and \textbf{C} display confusion matrices for target class sample size per modality of 100 and 10 respectively, whereas panel \textbf{B} shows the confusion matrix under the missing modality condition.}
    \label{Fig6}
\end{figure}

The TCGA dataset exhibits inherent class imbalance. In contrast to MODIS, MIDAS presented low stability when increasing the number of smaller classes (Fig.~\ref{Fig5}B). Moreover its inherent requirement of a mosaic structure in the training dataset made it difficult to experiment further in the high class imbalance setting where only a few samples are available in a given class or missing modality/class pair setting. We could however investigate the effect of imbalance in an extreme scenario for our model MODIS. We systematically varied the number of samples per modality in the smallest class of the 10-class subset, corresponding to Kidney Renal Clear Cell Carcinoma (KIRC) (Fig.~\ref{Fig6}A). Notably, with 50 samples per modality, MODIS recovered over 80\% recall for KIRC (data not shown) and achieved a global balanced accuracy above 92\% on the test set. Increasing to 100 samples yielded performance similar to that of the full dataset (Fig.~\ref{supp_tcga_2}A-B). These results underscore the robustness of our approach in handling severe class imbalance in high dimensional data.

We finally evaluated the case of a missing modality-class pair (Fig.~\ref{Fig6}B). In this setting, overall accuracy decreased by $\sim$4\% from the peak performance of 98\% on the full test set. The decrease was mainly caused by poor alignment in the latent space of a single modality (miRNA expression), which caused almost total loss of predictive power for this modality (Fig.~\ref{supp_tcga_2}C). In contrast, the modality-specific recall for all other modalities remained stable between 87–96\% across replicates. We tested the effect of introducing a small number of samples into the missing class–modality pair (Fig.~\ref{Fig6}C). With only 10 samples, the modality-specific recall recovered to an average of 66\% on the test set, with some replicates exceeding 80\% (Fig.~\ref{supp_tcga_2}D). Increasing to 20 samples further improved the recall to 76.4\%, and with 50 samples, performance nearly matched the full dataset, reaching the overall peak balanced accuracy of 94.8\% and an average modality-specific recall of 89.6\% for miRNA expression.

\section{Discussion}
We introduced MODIS, a framework designed to address one of the major challenges in multi-omics research: how to perform robust data integration when samples are scarce, unpaired across modalities, and only partially annotated. By combining coupled variational autoencoders with adversarial alignment and semi-supervised training, MODIS provides a stable architecture to learn a shared latent representation under these highly constrained conditions. Our results show that MODIS consistently achieves accurate classification, faithful cross-modal translation, and resilience to imbalance, even when supervision is extremely limited. A key advance of MODIS is the explicit reframing of integration with small datasets as an imbalanced classification problem. By training simultaneously on a large reference dataset and a small target dataset, MODIS effectively leverages shared structure to stabilize the representation of rare classes. This strategy enables accurate predictions with as little as one labeled sample per class and modality - an especially relevant scenario for rare disease research for example, where collecting comprehensive multi-omics data remains impractical. Another strength of MODIS is its robustness to missing modalities. While the model cannot recover completely absent modality–class pairs, we show that the addition of only a handful of samples is sufficient to restore accurate predictions. This property is of practical importance in biomedical settings, where technical or logistical constraints often result in incomplete multi-modal datasets. Finally, we compared MODIS to MIDAS, an alternative multi-omics data integration approach. We found that MODIS outperfomed MIDAS for the task of predicting classes from the joint latent representation on the TCGA dataset. In contrast to MIDAS, MODIS favors clustered representation, a beneficial setting for efficient classification, which can be one of the reason to the difference in performance.

Despite these strengths, some limitations remain. First, unlike more interpretable models, such as matrix factorization methods \cite{hirst2025motl}, deep autoencoders operate as black boxes and like other neural network architectures, MODIS inherits a lack of interpretability. The latent representation of the VAEs is well suited for classification and translation, but the biological meaning of individual latent dimensions is difficult to trace. While the network successfully learns to align the latent representations between modalities, the extent to which individual latent dimensions correspond to meaningful features and how latent representations encode shared and modality-specific features remains unclear. This lack of interpretability poses challenges in biomedical applications where explainability is critical \cite{rajpurkar2022ai}. Potential solutions include post-hoc analysis techniques such as quantification of feature importance to improve our understanding of the model’s internal representations \cite{lundberg2017unified}. Such techniques are compatible with MODIS and can be directly implemented in subsequent studies. Alternatively, one could turn to biologically informed variational autoencoders \cite{doncevic2023biologically,selby2025beyond}. Second, although MODIS achieves high accuracy on both simulated and TCGA benchmarks, its scalability to even larger, more heterogeneous datasets remains to be fully assessed. In this regard, a key aspect to guarantee the applicability of such strategy would certainly include a characterization of the distance between the reference and the target dataset, and the evolution of the model accuracy with respect to it. 

Beyond multi-omics, the methodological contributions of MODIS align with broader themes in multimodal machine learning, including cross-modal translation, semi-supervised alignment, and learning under imbalance. These challenges are common in diverse domains such as vision–language models, audio–text alignment, or video understanding, where datasets are often incomplete or weakly annotated \cite{radford2021learning, elizalde2023clap}. By successfully addressing them in the demanding context of biomedical data, MODIS contributes to the development of general strategies for multimodal data integration. The growing abundance of generated biomedical data let us anticipate that we will be able to use more complex architectures, such as foundation models to solve the question of multi-omics data integration for small and unpaired datasets \cite{maniparambil2025harnessing, khan2023contrastive, merullo2022linearly, groger2025limited, kalfon2025scprint}. It is however of interest to note that (variational) autoencoders, as used in MODIS, remain the most efficient architecture for gene expression reconstruction from a latent representation and that simple tools such as PCA can outperform foundation models for the representation of gene expression \cite{fu2026benchmarking}.

\section*{Acknowledgments}

This research was supported by l’Agence Nationale de la Recherche (ANR), project ANR21-CE45-0001-01 and ANR-21-CE13-0007.

This work was granted access to the HPC resources of IDRIS under the allocation 2025-AD010316676 made by GENCI and the Core Cluster of the Institut Français de Bioinformatique (IFB).

\bibliographystyle{unsrt}
\bibliography{references.bib}

\clearpage

\appendix
\setcounter{figure}{0}
\renewcommand{\thefigure}{S\arabic{figure}}

\setcounter{table}{0}
\renewcommand{\thetable}{S\arabic{table}}

\section{Supplementary Materials}
\begin{table}[h!]
\centering
\verysmall
\begin{tabular}{l | c c c c | c c c c}
\hline
\makecell{\textbf{Cancer} \\ \textbf{type}} & \multicolumn{4}{c|}{\textbf{Train dataset}} & \multicolumn{4}{c}{\textbf{Test dataset}} \\
\cline{2-5}
\cline{6-9}
& \makecell{\textbf{DNA} \\ \textbf{methylation}} & \makecell{\textbf{Gene} \\ \textbf{expression}} & \makecell{\textbf{miRNA} \\ \textbf{expre}} & \textbf{Total} & \makecell{\textbf{DNA} \\ \textbf{methylation}} & \makecell{\textbf{Gene} \\ \textbf{expression}} & \makecell{\textbf{miRNA} \\ \textbf{expre}} & \textbf{Total} \\
\hline
BRCA & 615 & 875 & 861 & \textbf{2,363} & 157 & 219 & 216 & \textbf{592} \\
Normal & 587 & 579 & 528 & \textbf{1,694} & 147 & 144 & 131 & \textbf{422} \\
HNSC & 422 & 416 & 418 & \textbf{1,256} & 106 & 104 & 105 & \textbf{315} \\
LGG & 413 & 414 & 410 & \textbf{1,237} & 103 & 102 & 102 & \textbf{307} \\
THCA & 406 & 404 & 405 & \textbf{1,215} & 101 & 101 & 101 & \textbf{303} \\
UCEC & 346 & 436 & 431 & \textbf{1,213} & 85 & 109 & 107 & \textbf{301} \\
PRAD & 399 & 398 & 395 & \textbf{1,192} & 99 & 99 & 99 & \textbf{297} \\
LUAD & 367 & 414 & 411 & \textbf{1,192} & 91 & 102 & 102 & \textbf{295} \\
SKCM & 376 & 375 & 359 & \textbf{1,110} & 94 & 94 & 89 & \textbf{277} \\
KIRC & 256 & 427 & 413 & \textbf{1,096} & 63 & 106 & 103 & \textbf{272} \\
LUSC & 296 & 401 & 382 & \textbf{1,079} & 74 & 100 & 96 & \textbf{270} \\
STAD & 315 & 329 & 348 & \textbf{992} & 80 & 83 & 88 & \textbf{251} \\
BLCA & 329 & 325 & 327 & \textbf{981} & 83 & 81 & 82 & \textbf{246} \\
COAD & 236 & 366 & 355 & \textbf{957} & 59 & 92 & 89 & \textbf{240} \\
LIHC & 302 & 297 & 298 & \textbf{897} & 75 & 74 & 74 & \textbf{223} \\
CESC & 246 & 244 & 246 & \textbf{736} & 61 & 60 & 61 & \textbf{182} \\
OV & 8 & 338 & 389 & \textbf{735} & 2 & 84 & 97 & \textbf{183} \\
KIRP & 220 & 232 & 233 & \textbf{685} & 55 & 58 & 58 & \textbf{171} \\
SARC & 209 & 208 & 207 & \textbf{624} & 52 & 51 & 52 & \textbf{155} \\
ESCA & 149 & 148 & 148 & \textbf{445} & 36 & 36 & 36 & \textbf{108} \\
PCPG & 144 & 144 & 144 & \textbf{432} & 35 & 35 & 35 & \textbf{105} \\
PAAD & 147 & 142 & 142 & \textbf{431} & 37 & 36 & 36 & \textbf{109} \\
LAML & 155 & 121 & 150 & \textbf{426} & 39 & 30 & 38 & \textbf{107} \\
TGCT & 120 & 120 & 120 & \textbf{360} & 30 & 30 & 30 & \textbf{90} \\
READ & 79 & 133 & 129 & \textbf{341} & 19 & 33 & 32 & \textbf{84} \\
THYM & 99 & 96 & 99 & \textbf{294} & 25 & 24 & 25 & \textbf{74} \\
GBM & 111 & 129 & 6 & \textbf{246} & 28 & 33 & 2 & \textbf{63} \\
MESO & 70 & 70 & 70 & \textbf{210} & 17 & 17 & 17 & \textbf{51} \\
UVM & 64 & 64 & 64 & \textbf{192} & 16 & 16 & 16 & \textbf{48} \\
ACC & 64 & 63 & 64 & \textbf{191} & 16 & 16 & 16 & \textbf{48} \\
KICH & 53 & 53 & 53 & \textbf{159} & 13 & 13 & 13 & \textbf{39} \\
UCS & 46 & 46 & 46 & \textbf{138} & 11 & 11 & 11 & \textbf{33} \\
DLBC & 39 & 39 & 38 & \textbf{116} & 9 & 9 & 9 & \textbf{27} \\
CHOL & 29 & 28 & 29 & \textbf{86} & 7 & 7 & 7 & \textbf{21} \\
\hline
\multicolumn{1}{l}{\textbf{Total}} & \textbf{7,729} & \textbf{8,874} & \textbf{8,718} & \textbf{25,321} & \textbf{1,925} & \textbf{2,209} & \textbf{2,175} & \textbf{6,309} \\
\hline
\end{tabular}
\caption{\textbf{TCGA multi-omics dataset}. Number of samples in the training and test sets for each modality. These counts are consistent across all classes in the subsets detailed in Fig.~\ref{Fig5}.}
\label{tab:tcga_data}
\end{table}

\begin{table}[h!]
\centering
\verysmall
\begin{tabular}{l | c c c c | c c c c}
\hline
\makecell{\textbf{Cancer} \\ \textbf{type}} & \multicolumn{4}{c|}{\textbf{Train dataset}} & \multicolumn{4}{c}{\textbf{Test dataset}} \\
\cline{2-5}
\cline{6-9}
& \makecell{\textbf{DNA} \\ \textbf{methylation}} & \makecell{\textbf{Gene} \\ \textbf{expression}} & \makecell{\textbf{miRNA} \\ \textbf{expre}} & \textbf{Total} & \makecell{\textbf{DNA} \\ \textbf{methylation}} & \makecell{\textbf{Gene} \\ \textbf{expression}} & \makecell{\textbf{miRNA} \\ \textbf{expre}} & \textbf{Total} \\
\hline
BRCA & 593 & 593 & 593 & \textbf{1,779} & 165 & 165 & 165 & \textbf{495} \\
Normal & 295 & 295 & 295 & \textbf{885} & 61 & 61 & 61 & \textbf{183} \\
HNSC & 416 & 416 & 416 & \textbf{1,248} & 97 & 97 & 97 & \textbf{291} \\
LGG & 406 & 406 & 406 & \textbf{1,218} & 94 & 94 & 94 & \textbf{282} \\
THCA & 392 & 392 & 392 & \textbf{1,176} & 108 & 108 & 108 & \textbf{324} \\
UCEC & 341 & 341 & 341 & \textbf{1,023} & 87 & 87 & 87 & \textbf{261} \\
PRAD & 366 & 366 & 366 & \textbf{1,098} & 123 & 123 & 123 & \textbf{369} \\
LUAD & 361 & 361 & 361 & \textbf{1,083} & 85 & 85 & 85 & \textbf{255} \\
SKCM & 363 & 363 & 363 & \textbf{1,089} & 83 & 83 & 83 & \textbf{249} \\
KIRC & 252 & 252 & 252 & \textbf{756} & 62 & 62 & 62 & \textbf{186} \\
LUSC & 295 & 295 & 295 & \textbf{885} & 70 & 70 & 70 & \textbf{210} \\
STAD & 295 & 295 & 295 & \textbf{885} & 75 & 75 & 75 & \textbf{225} \\
BLCA & 317 & 317 & 317 & \textbf{951} & 84 & 84 & 84 & \textbf{252} \\
COAD & 232 & 232 & 232 & \textbf{696} & 51 & 51 & 51 & \textbf{153} \\
LIHC & 300 & 300 & 300 & \textbf{900} & 65 & 65 & 65 & \textbf{195} \\
CESC & 241 & 241 & 241 & \textbf{723} & 62 & 62 & 62 & \textbf{186} \\
OV & 7 & 7 & 7 & \textbf{21} & 2 & 2 & 2 & \textbf{6} \\
KIRP & 212 & 212 & 212 & \textbf{636} & 61 & 61 & 61 & \textbf{183} \\
SARC & 189 & 189 & 189 & \textbf{567} & 66 & 66 & 66 & \textbf{198} \\
ESCA & 149 & 149 & 149 & \textbf{447} & 33 & 33 & 33 & \textbf{99} \\
PCPG & 145 & 145 & 145 & \textbf{435} & 31 & 31 & 31 & \textbf{93} \\
PAAD & 150 & 150 & 150 & \textbf{450} & 27 & 27 & 27 & \textbf{81} \\
LAML & 112 & 112 & 112 & \textbf{336} & 22 & 22 & 22 & \textbf{66} \\
TGCT & 116 & 116 & 116 & \textbf{348} & 29 & 29 & 29 & \textbf{87} \\
READ & 81 & 81 & 81 & \textbf{243} & 16 & 16 & 16 & \textbf{48} \\
THYM & 93 & 93 & 93 & \textbf{279} & 27 & 27 & 27 & \textbf{81} \\
GBM & 2 & 2 & 2 & \textbf{6} & 1 & 1 & 1 & \textbf{3} \\
MESO & 70 & 70 & 70 & \textbf{210} & 17 & 17 & 17 & \textbf{51} \\
UVM & 65 & 65 & 65 & \textbf{195} & 15 & 15 & 15 & \textbf{45} \\
ACC & 66 & 66 & 66 & \textbf{198} & 13 & 13 & 13 & \textbf{39} \\
KICH & 51 & 51 & 51 & \textbf{153} & 15 & 15 & 15 & \textbf{45} \\
UCS & 44 & 44 & 44 & \textbf{132} & 13 & 13 & 13 & \textbf{39} \\
DLBC & 41 & 41 & 41 & \textbf{123} & 6 & 6 & 6 & \textbf{18} \\
CHOL & 29 & 29 & 29 & \textbf{87} & 6 & 6 & 6 & \textbf{18} \\
\hline
\multicolumn{1}{l}{\textbf{Total}} & \textbf{7,087} & \textbf{7,087} & \textbf{7,087} & \textbf{21,261} & \textbf{1,772} & \textbf{1,772} & \textbf{1,772} & \textbf{5,316} \\
\hline
\end{tabular}
\caption{\textbf{TCGA multi-omics dataset (paired samples only)}. Number of samples in the training and test sets for each modality. These counts represent the strictly intersection-paired patient entries across all classes. The paired samples are used as ground truth for the evaluation of the reconstruction error.}
\label{tab:tcga_data_paired}
\end{table}

\clearpage

\subsection{Network architecture}

\begin{figure}[!htp]
    \centering
    \includegraphics[width=0.75\textwidth]{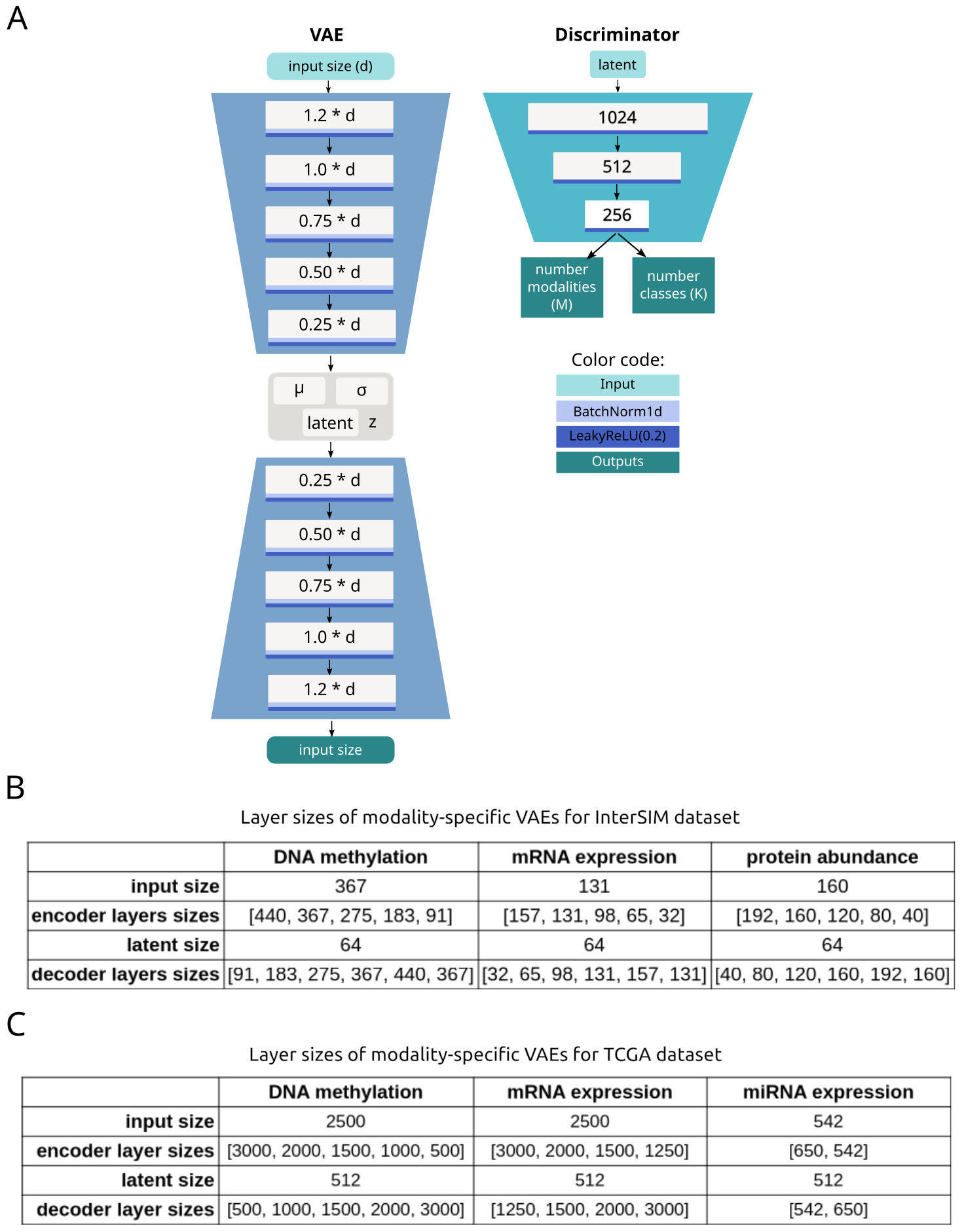}
    \caption{\textbf{Architecture of the MODIS model}. \textbf{A}, Default (adaptable) VAE structure, with layer sizes tailored to each modality. The discriminator is adapted from the AC-GAN design \cite{odena2017conditional}, with two output branches, one predicting modality and the other class labels. \textbf{B}, Modality-specific VAE layer sizes used for the InterSIM dataset. \textbf{C}, Modality-specific VAE layer sizes used for the TCGA dataset.}
    \label{supp_model}
\end{figure}

The architecture of MODIS is shown in Fig.~\ref{supp_model}, and the number of model parameters for each dataset is provided in Table~\ref{tab:modis_parameters}.

\begin{table}[h]
\centering
\begin{tabular}{lrr}
\hline
\textbf{Module} & \textbf{InterSIM} & \textbf{TCGA (10 classes)} \\
\hline
DNA Methylation VAE & 1,007,647 & 37,818,524 \\
RNA Expression VAE & 134,081 & 38,717,524 \\
Protein Abundance VAE / miRNA Expression VAE & 198,320 & 2,249,346 \\
Discriminator & 724,744 & 1,184,781 \\
\hline
\textbf{Total Parameters} & 2,064,792 & 79,970,175 \\
\hline
\end{tabular}
\caption{Number of parameters for MODIS architectures on InterSIM and TCGA (10 classes subset) datasets.}
\label{tab:modis_parameters}
\end{table}

\begin{table}[h]
\centering
\begin{tabular}{lcc}
\hline
\textbf{Parameter} & \textbf{InterSIM} & \textbf{TCGA} \\
\hline
Epochs & 300 & 4000 \\
Batch size & 32 & 32 \\
Learning rate & $1\times 10^{-4}$ & $1\times 10^{-4}$ \\
Beta & $1\times 10^{-4}$ & $1\times 10^{-6}$ \\
Beta1 & 0.5 & 0.5 \\
$\lambda_r$ & 10 & 160 \\
\hline
\end{tabular}
\caption{Training hyperparameters for the InterSIM and TCGA models.}
\label{tab:model_hyperparameters}
\end{table}

\begin{figure}[!htp]
    \centering
    \includegraphics[width=0.75\textwidth]{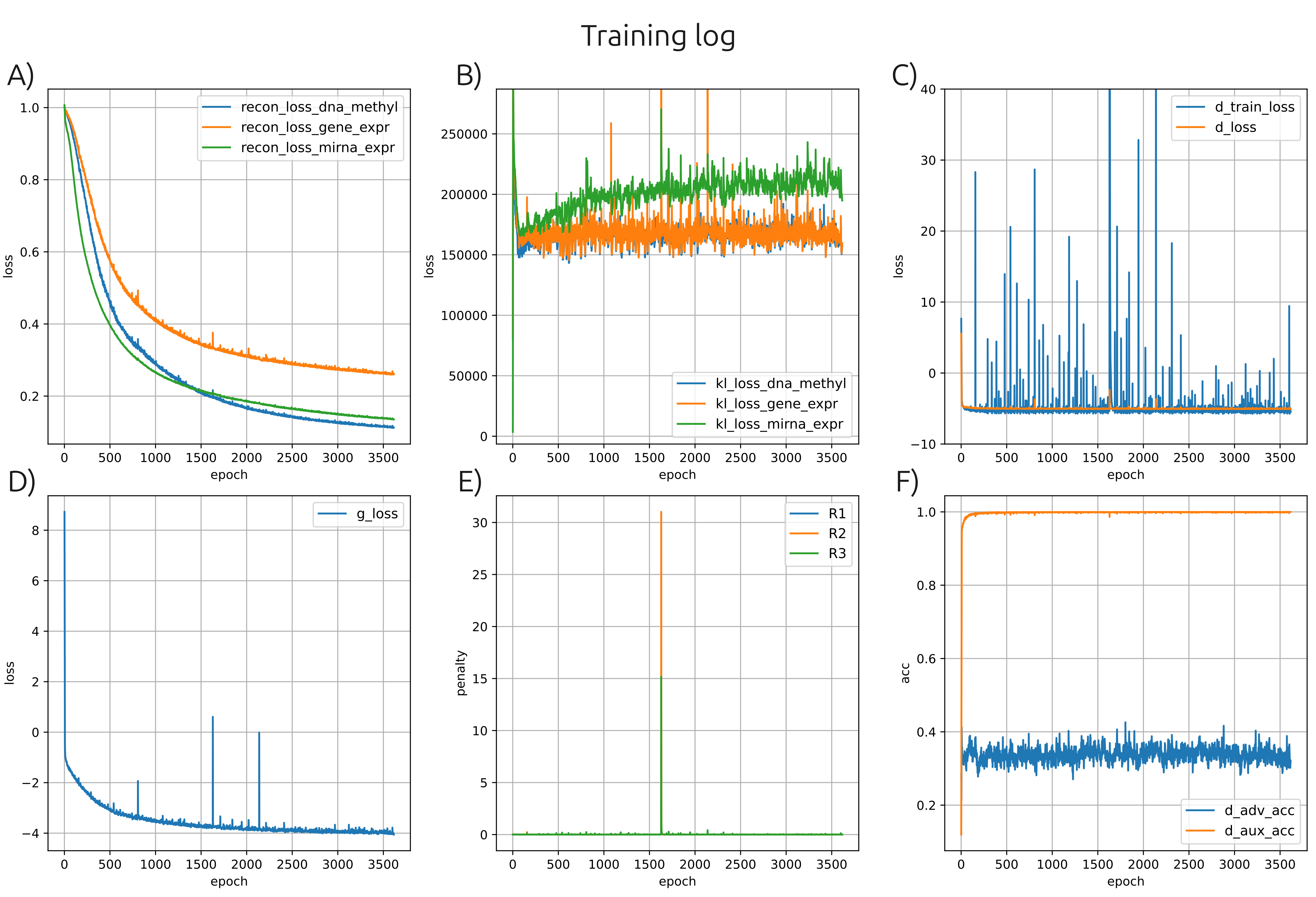}
    \caption{\textbf{Training curve of MODIS on TCGA 34 classes dataset}. \textbf{A}, Reconstruction loss for each of the modalities. \textbf{B}, KL Loss in the latent space, for each of the modalities. \textbf{C} Discriminator loss in the adversarial training setting. d\_train\_loss corresponds to the step where the discriminator is being trained and d\_loss is the evaluation of the discriminator loss when the VAE are being trained. \textbf{D} Global loss, a linear combination of all the specific loss functions. \textbf{E} Regularization penalty in the R3GAN loss for each modality. \textbf{F} Accuracy of the discriminator. d\_adv\_acc corresponds to the modality classification ($0.33$ is expected for three undistinguishable modalities) and d\_aux\_acc corresponds to the class classification.}
    \label{training_curves}
\end{figure}
s

\begin{figure}
    \centering
    \includegraphics[width=0.95\textwidth]{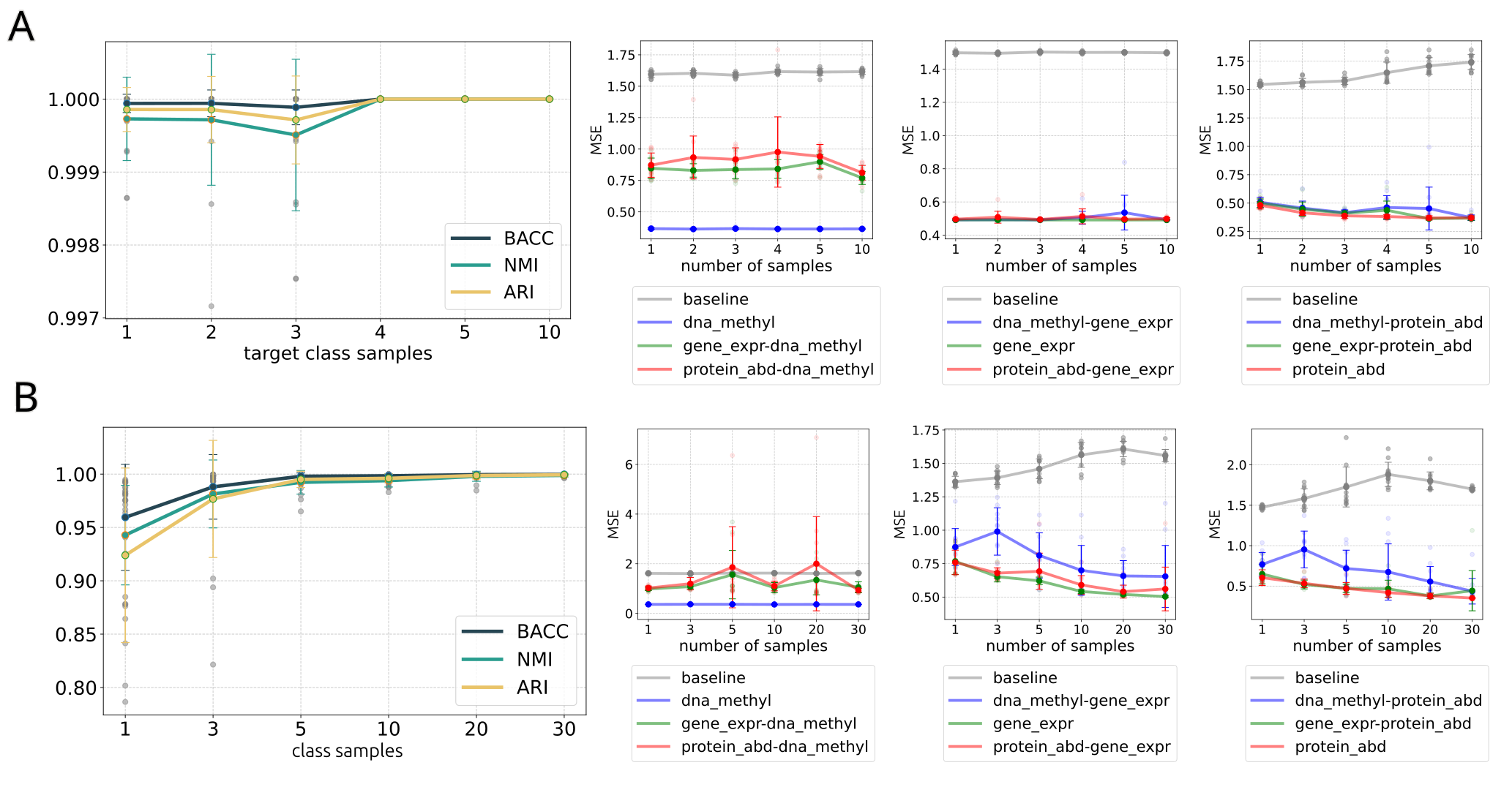}
    \caption{\textbf{Evaluation of the effect of partially missing modalities.} \textbf{A}, Setting with one partially missing class–modality pair (Fig.~\ref{Fig4} ii). \textbf{B}, Setting with two partially missing class–modality pairs (Fig.~\ref{Fig4} iv). Columns: (1) class prediction performance (balanced accuracy (BACC), normalized mutual information (NMI), and adjusted rand index (ARI)) as a function of the number of training samples in the corresponding class–modality pairs; (2–4) scatter plots of mean squared error (MSE, mean $\pm$ SD) for modality reconstruction and translation between MODIS predictions and ground truth, plotted against the number of target class samples. The baseline corresponds to the MSE across all sample pairs within each modality. Each data point represents the mean over 10 replicates.}
    \label{supp_eval_intersim}
\end{figure}

\begin{figure}
    \centering
    \includegraphics[width=0.95\textwidth]{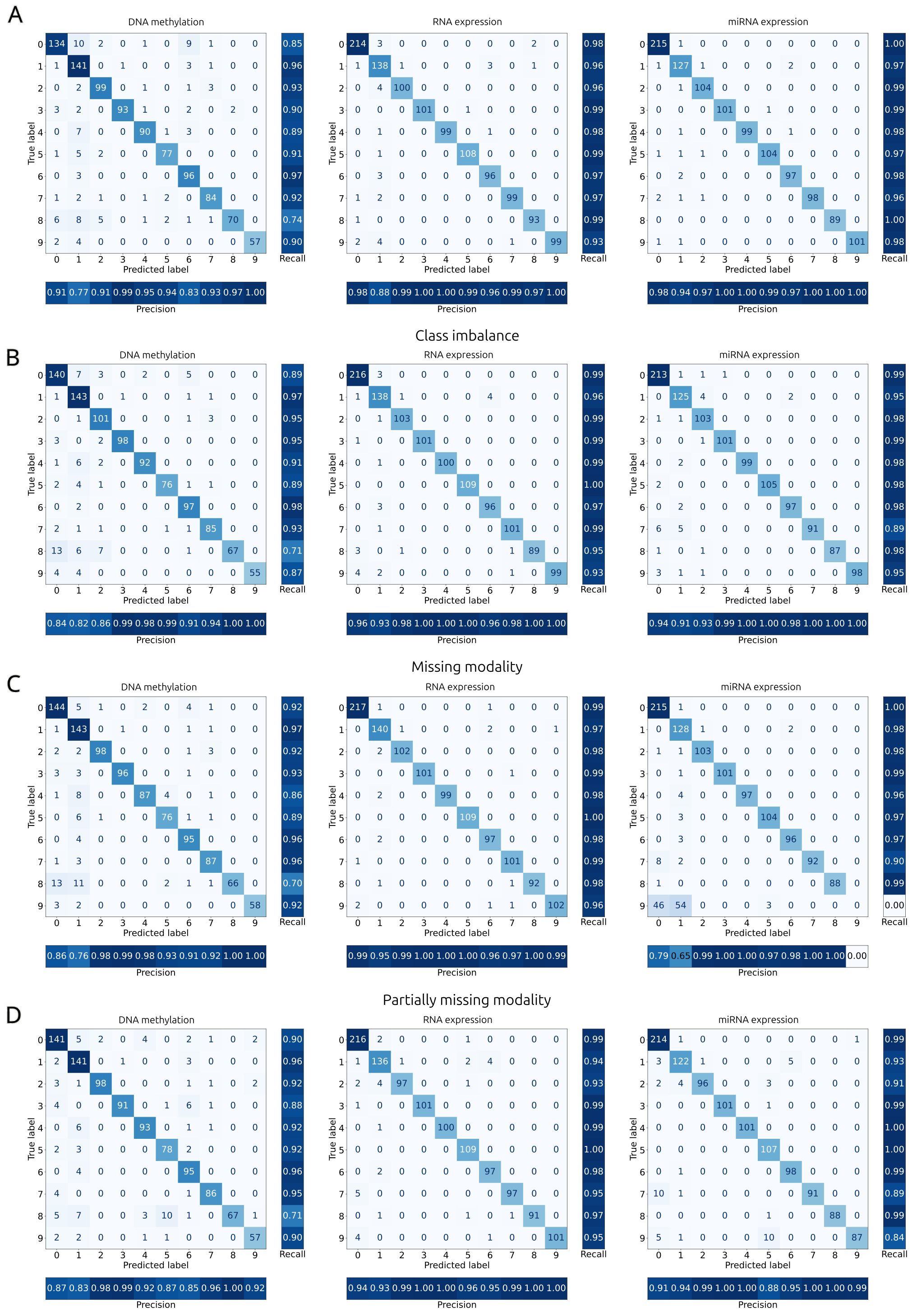}
    \caption{\textbf{Modality-wise confusion matrices from TCGA evaluations}. Columns correspond (left to right) to DNA methylation, RNA expression, and miRNA expression. Panel \textbf{A} shows the modality-wise confusion matrix of MODIS discriminator predictions on the test dataset corresponding to the TCGA 10 classes setting. Panels \textbf{B}–\textbf{D} show modality-wise confusion matrices corresponding to the confusion matrices of the overall predictions in Fig.~\ref{Fig6}, for the evaluations under class imbalance, missing modality, and partially missing modality, respectively.}
    \label{supp_tcga_2}
\end{figure}

\begin{figure}
    \centering
    \includegraphics[width=0.95\textwidth]{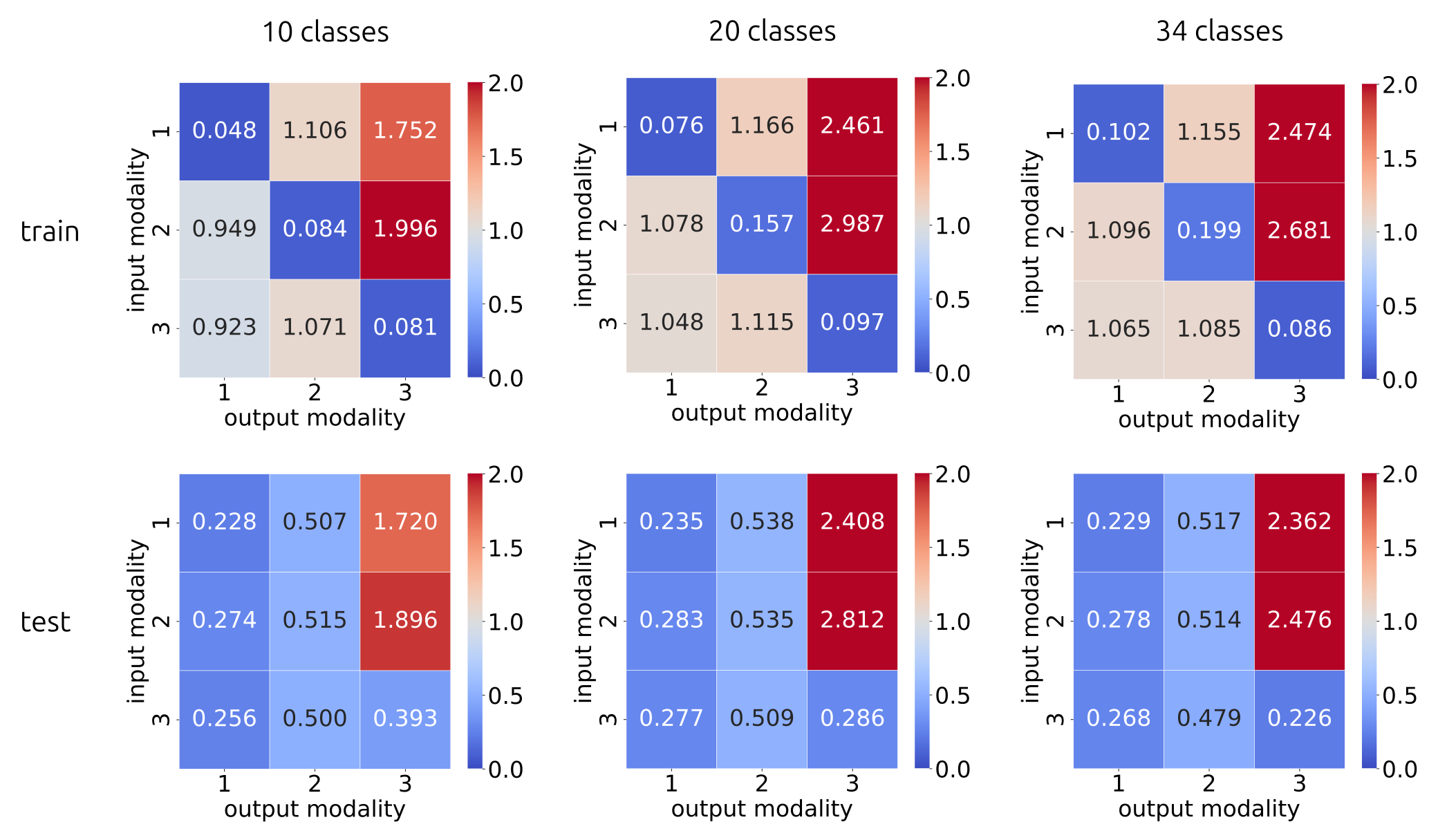}
    \caption{\textbf{MODIS Reconstruction Error on the TCGA dataset} Heatmaps with the MSE for the modality reconstruction and translation between MODIS prediction and ground truth for 10, 20 and 34 classes on the train and the test set (Modality 1: DNA Methylation, Modality 2: RNA expression, Modality 3: miRNA expression).}
    \label{MODIS_reconstruction_error}
\end{figure}
\clearpage

\subsection{MODIS training} \label{modis_training}

The complete MODIS architecture is reported on Fig. \ref{supp_model}. We describe in this section the training procedure. The training dataset is assumed to be composed of $n$ unpaired samples distributed across the $M$ modalities and the $K$ classes, which we denote by triplets $(x_i,m_i,c_i)_{i \in \{1, \ldots, n\}}$, where $x_i \in \mathbb{R}^{p_{m_i}}$ is sample $i$ with $p_{m_i}$ features, $m_i \in \{1, \ldots, M\}$ indicates the modality and $c_i \in \{1, \ldots, K\}$ indicates the class of the given sample.

MODIS is composed of $M$ variational autoencoders, one for each modality, all having the same latent dimension. For modality $m$, we denote by $E^{(m)}$ and $D^{(m)}$ the encoder and decoder, composed of multiple fully connected layers and non-linearities (see Fig. \ref{supp_model}). For a sample $i$, we denote by $z_i \sim \mathcal{N}(\mu_i, \sigma_i)$ the latent representation obtained by encoder $(\mu_i, \sigma_i) = E^{(m_i)}(x_i)$, and we denote by $\hat{x}_i = D^{(m_i)}(z_i)$ the reconstructed samples obtained by decoder $D^{(m_i)}$, and by $\hat{x}_i^{(m_i, m_b)} = D^{(m_b)}(z_i)$, the translation obtained by combining encoder $E^{(m_i)}$ with decoder $D^{(m_b)}$ associated to modality $m_b \neq m_i$.

To train the variational autoencoders, we relied on classical reconstruction loss $L_{\text{recon}}$ and Kullback-Leibler regularization \cite{kingma2013auto}.

\begin{equation*}
    L_{\text{recon}}= \frac{1}{n} \sum_{i=1}^{n} ||x_i - \hat{x}_i||_2^2.
\end{equation*}

In practice, $L_{\text{recon}}^{(m)}$ is computed separately for each modality specific VAEs and summed in a global reconstruction loss $L_{\text{recon}}$.

To ensure regularity in the latent space, we use the Kullback-Leibler divergence:
\begin{equation*}
    D_{KL} = \frac{1}{n} \sum_{i = 1}^{n} KL(\mathcal{N}(\mu_i, \sigma_i), \mathcal{N}(0,1)).
\end{equation*}
As for the reconstruction loss, in practice, $D_{KL}^{(m)}$ is computed separately for each modality specific VAEs and summed in a global Kullback-Leibler divergence $D_{KL}$.

In addition to the variational autoencoders, MODIS is composed of a discriminator $D$ acting on the latent representation. The discriminator $D$ has two output branches, first $D_{\text{mod}}$ having dimension $M$ and aiming at predicting the modality from the latent representation $z$, and $D_{\text{class}}$ having dimension $K$ and aiming at predicting the class $c$ from the latent representation. The discriminator $D$ is composed of three fully connected layer, with LeakyReLU non-linear function, followed by two distinct fully connected layers leading to $D_{\text{mod}}$ and $D_{\text{class}}$ respectively, see Fig. \ref{supp_model}. 

The encoders of the VAEs and $D_{\text{mod}}$ are trained in an adversarial way such that the latent representation becomes "modality free"\cite{DBLP:journals/corr/MakhzaniSJG15}. Following \cite{huang2024gan}, to ensure stability during training, we adapted the regularized relativistic GAN loss (R3GAN) to $M$ modalities in the following way.

In the training step, when the discriminator $D_{\text{mod}}$ is trained to identify modality from the latent representation, we minimize the following loss function:

\begin{multline*}
L_{\text{relativistic},D_{\text{mod}}} = \mathbb{E}_{\substack{x^{(1)}\sim P_1\\ \vdots \\ x^{(M)}\sim P_M}}\left[ f \left( -\sum_{i = 1}^M \left( D_{{\text{mod}},\text{logits}}(x^{(i)}) - \sum_{\substack{j = 1 \\ j \neq i}}^M D_{{\text{mod}},\text{logits}}(x^{(j)}) \right) \right) \right]\\+ \frac{\gamma}{2} \sum_{i = 1}^M \mathbb{E}_{x^{(i)} \sim Pi} \left[ || \nabla_{x^{(i)}} D_{{\text{mod}},\text{logits}}(x^{(i)}) ||^2   \right]
\end{multline*}
with $f$ being the softplus function ($f(x) = \ln(1 + e^x)$) and $D_{{\text{mod}},\text{logits}}(x^{(i)})$ denotes the logits output of the modality discriminator applied on the latent representation obtained from sample $x^{(i)}$, drawn from the distribution of points in modality $i$ (noted as $Pi$).

In the training step, when the encoders of the VAEs are trained to fool the discriminator $D_{\text{mod}}$, we minimize the following adversarial loss function:

\begin{equation*}
L_{\text{relativistic},\text{VAE}} = \mathbb{E}_{\substack{x^{(1)}\sim P_1\\ \vdots \\ x^{(M)}\sim P_M}}\left[ f \left( -\sum_{i = 1}^M \left( \sum_{\substack{j = 1 \\ j \neq i}}^M \left(D_{{\text{mod}},\text{logits}}(x^{(j)}) - D_{{\text{mod}},\text{logits}}(x^{(i)}) \right) \right) \right) \right].
\end{equation*}

To train the class classifier $D_{\text{class}}$, we minimize the classical cross-entropy loss:
\begin{equation*}
L_C = - \frac{1}{N}\sum_{i = 1}^N\sum_{k = 1}^{K} c_i\log(\hat{c}_{i,k})
\end{equation*}
where $c_i$ is the true label of sample $i$ (0 or 1 for each class) and $\hat{c}_{i,k}$ is the predicted probability for class $k$ and sample $i$.

Moreover, to ensure that the latent representation has good cluster separability, we adapted a clustering loss $L_{\text{clustering}}$ used in \cite{tang2023explainable} and initially proposed in \cite{kampffmeyer2019deep} which is composed of three terms $L_{\text{clustering}} = L_{C1} + L_{C2} + L_{C3}$.

The first term aims at separating as much as possible cluster soft assignments from the various classes.

\begin{equation*}
L_{C1} = \binom{K}{2}^{-1} \sum_{i = 1}^{K-1}\sum_{j > i}^{K} \frac{\tilde{\alpha}_i^{T}S\tilde{\alpha}_j}{\sqrt{\tilde{\alpha}_i^{T}S\tilde{\alpha}_i}\sqrt{\tilde{\alpha}_j^{T}S\tilde{\alpha}_j}}
\end{equation*}
where $\tilde{\alpha}_i$ is the class soft assignment and the matrix $S = (s_{i,j})_{\{1 .. n\}\times \{1 .. n\}}$ is defined by $s_{i,j} = \exp{-\frac{||h_i - h_j||^2_2}{2 \sigma^2}}$, where $h$ is the last layer of the discriminator $D$ before branching into $D_{\text{class}}$ with a fully connected layer.

The second term pushes the soft assignment values of the different classes to distinct corners of a simplex in $\mathbb{R}^K$. Let's denote by $e_k \in \mathbb{R}^K$ a corner of the simplex, it's a one hot encoding with value 1 on the $k$-th component. $\alpha_i$ is the soft assignment of sample $i$ among the $K$ classes, and $p_k \in \mathbb{R}^n$ where its $j$-th component is $\exp( - || \alpha_j - e_k||_2^2)$. The second term of the loss reads.
\begin{equation*}
L_{C2} = \binom{K}{2}^{-1}\sum_{i = 1}^{K-1}\sum_{j > i}^{K} \frac{p_i^{T}Sp_j}{\sqrt{p_i^{T}Sp_i}\sqrt{p_j^{T}Sp_j}}.
\end{equation*}

The third term aims to promote diversity in the prediction and avoid trivial solutions where most of the samples are assigned to a small subset of the labels, using negative entropy. We define $\overline{\alpha} \in \mathbb{R}^K$ as the average of the $\alpha_i$ for $i \in \{1, \ldots, n\}$, and with $k$-th element $\overline{\alpha}_k$.
\begin{equation*}
L_{C3} = \sum_{k = 1}^{K}\overline{\alpha}_k \log \overline{\alpha}_k
\end{equation*}

As explained in the Methods section, MODIS in trained on a dataset with $n$ samples, through two alternative steps. The first step concerns the update of the discriminator's parameters, which is obtained by minimizing the loss $L_D$ (equation \ref{eqn_D}), while the VAEs parameters are frozen. The second step concerns the update of the VAEs parameters, obtained by minimizing the loss $L_{VAE}$ (equation \ref{eqn_VAE}), while the discriminator's parameters are frozen. We reproduce here $L_D$ and $L_{VAE}$ and provide below a description of the various terms:

\begin{equation*}
   L_D = L_{\text{relativistic},D} + L_C + L_{\text{clustering}}
\end{equation*}

\begin{equation*}
   L_{VAE} =  \sum_{i = 1}^{M}\left(L_{\text{recon}}^{(i)}+ \beta D_{KL}^{(i)}\right)+ L_{\text{relativistic},\text{VAE}} + L_C + L_{\text{clustering}} 
\end{equation*}

In the semi-supervised setting, when only a few training samples have known labels, we compute the classification loss $L_C$ only on the labeled samples, the other aspects of the training scheme remain the same.

\newpage

\begin{figure}
    \centering
    \includegraphics[width=0.95\textwidth]{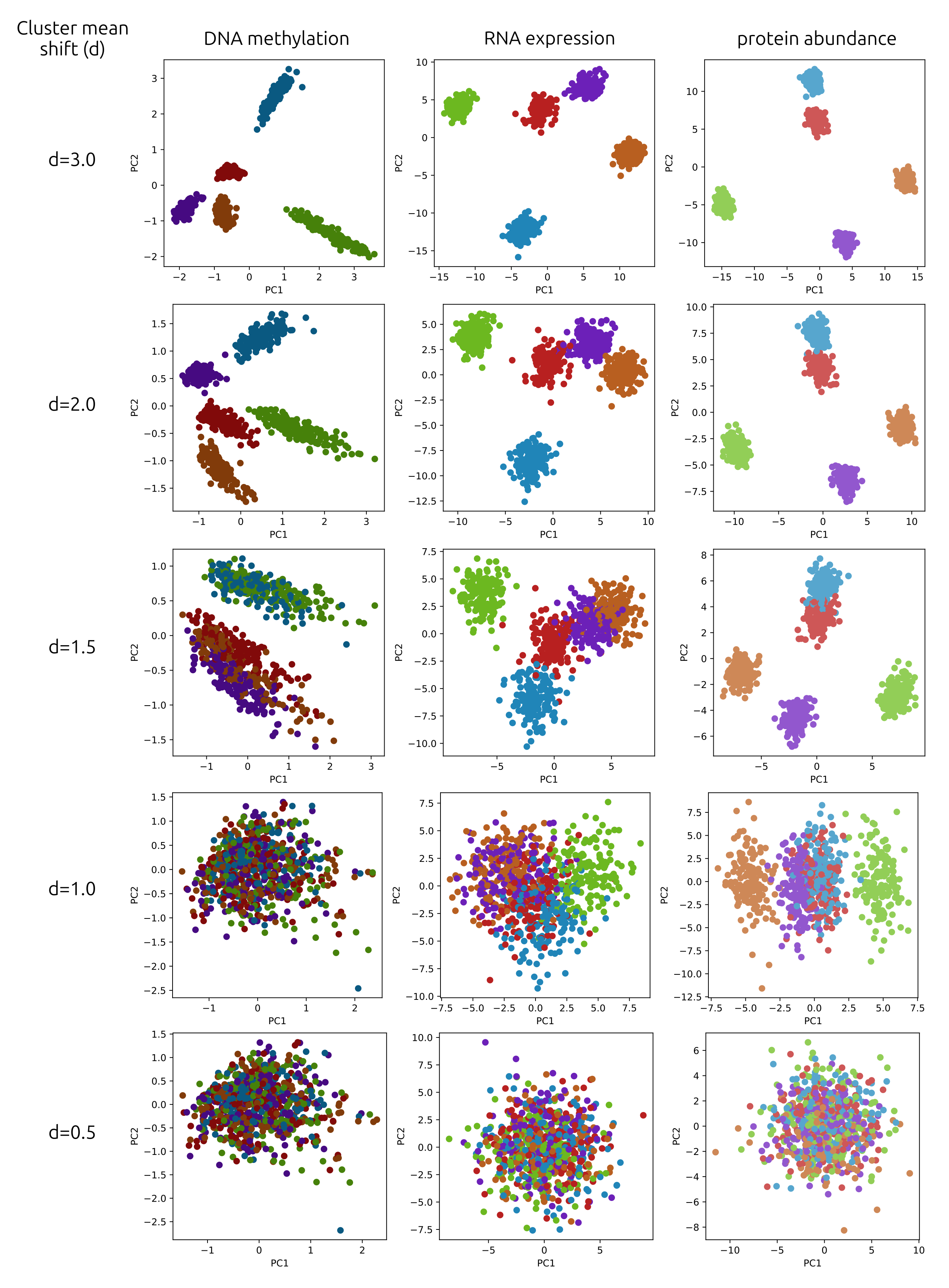}
    \caption{\textbf{INTERSIM datasets} Two-dimensional PCA representation of each modality of INTERSIM datasets generated with various values of the cluster mean shift parameter (d).}
    \label{intersim_datasets}
\end{figure}

\begin{figure}
    \centering
    \includegraphics[width=0.95\textwidth]{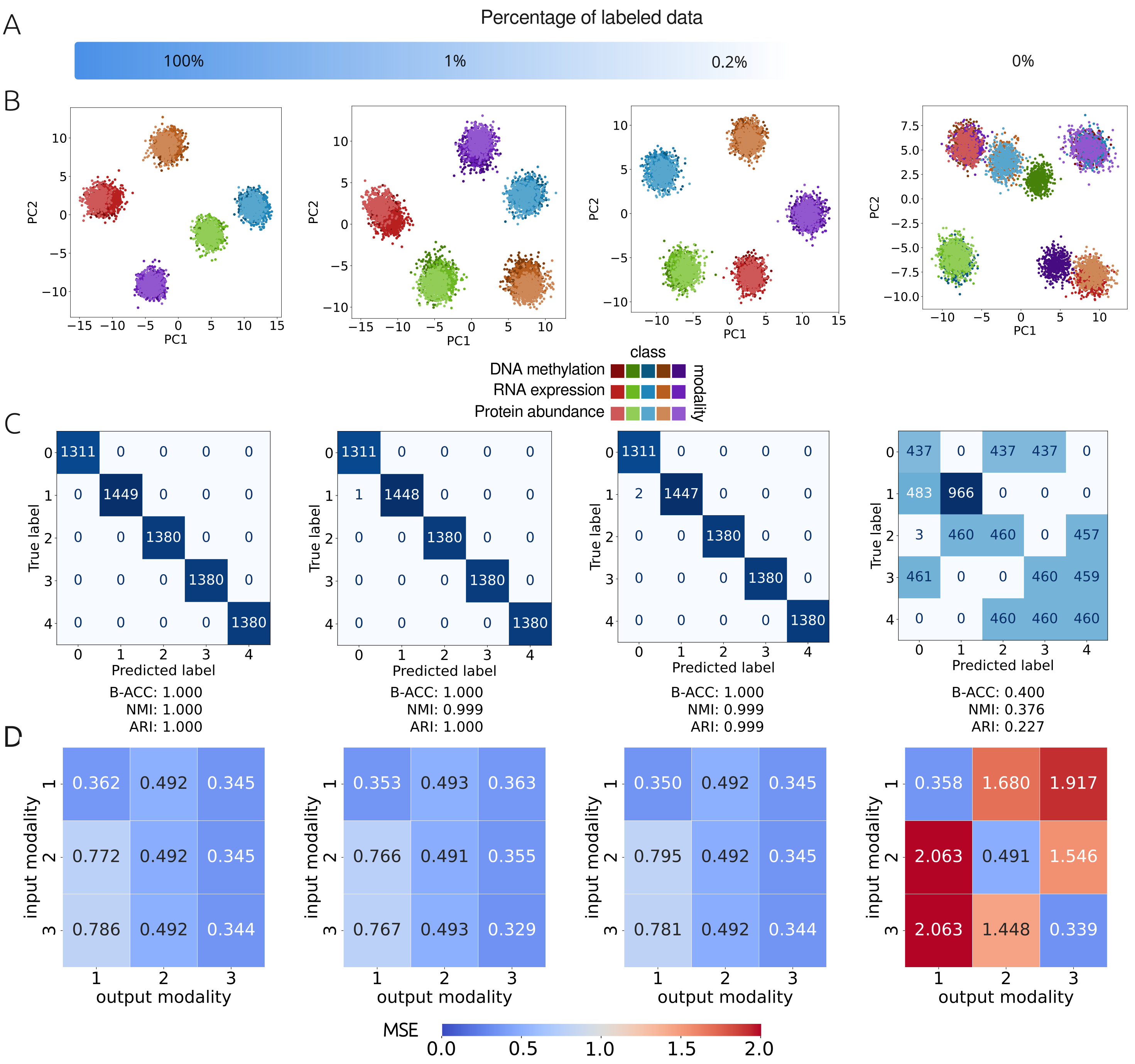}
    \caption{Latent Alignment obtained with MODIS - INTERSIM dataset with cluster mean shift parameter d = 2.0}
    \label{intersim_2.0d}
\end{figure}

\begin{figure}
    \centering
    \includegraphics[width=0.95\textwidth]{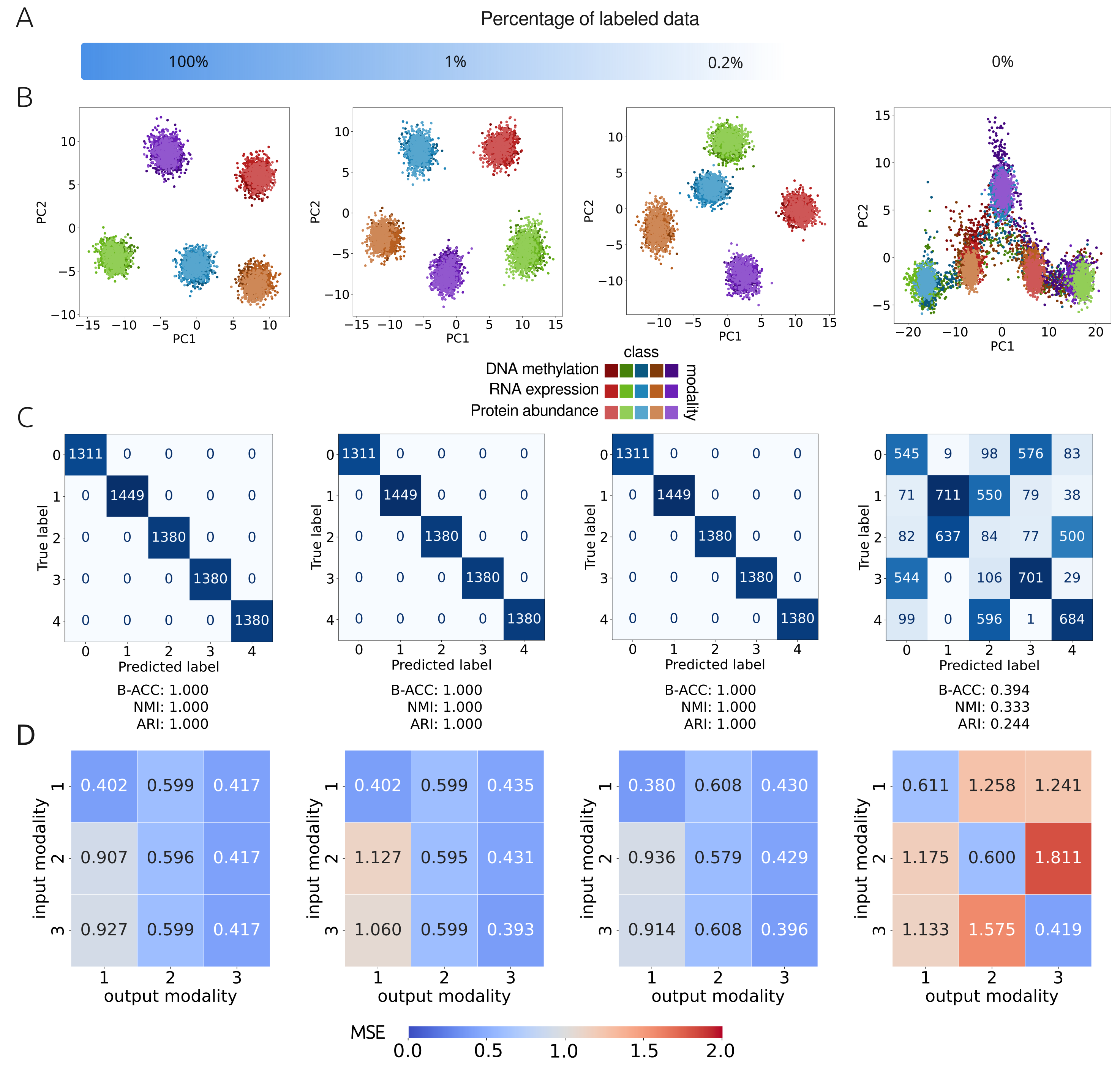}
    \caption{Latent Alignment obtained with MODIS - INTERSIM dataset with cluster mean shift parameter d = 1.5}
    \label{intersim_1.5d}
\end{figure}

\begin{figure}
    \centering
    \includegraphics[width=0.95\textwidth]{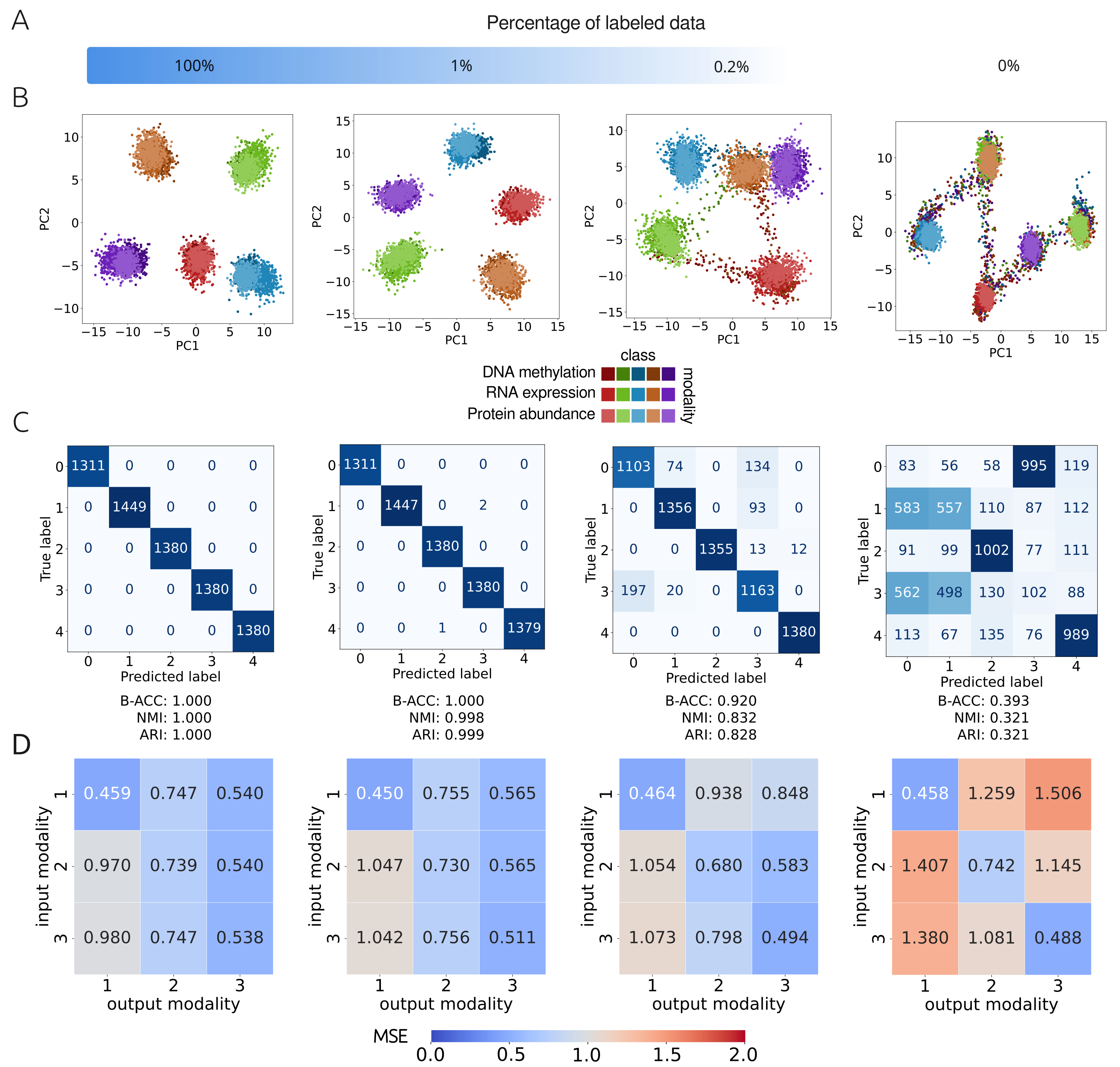}
    \caption{Latent Alignment obtained with MODIS - INTERSIM dataset with cluster mean shift parameter d = 1.0}
    \label{intersim_1.0d}
\end{figure}

\begin{figure}
    \centering
    \includegraphics[width=0.95\textwidth]{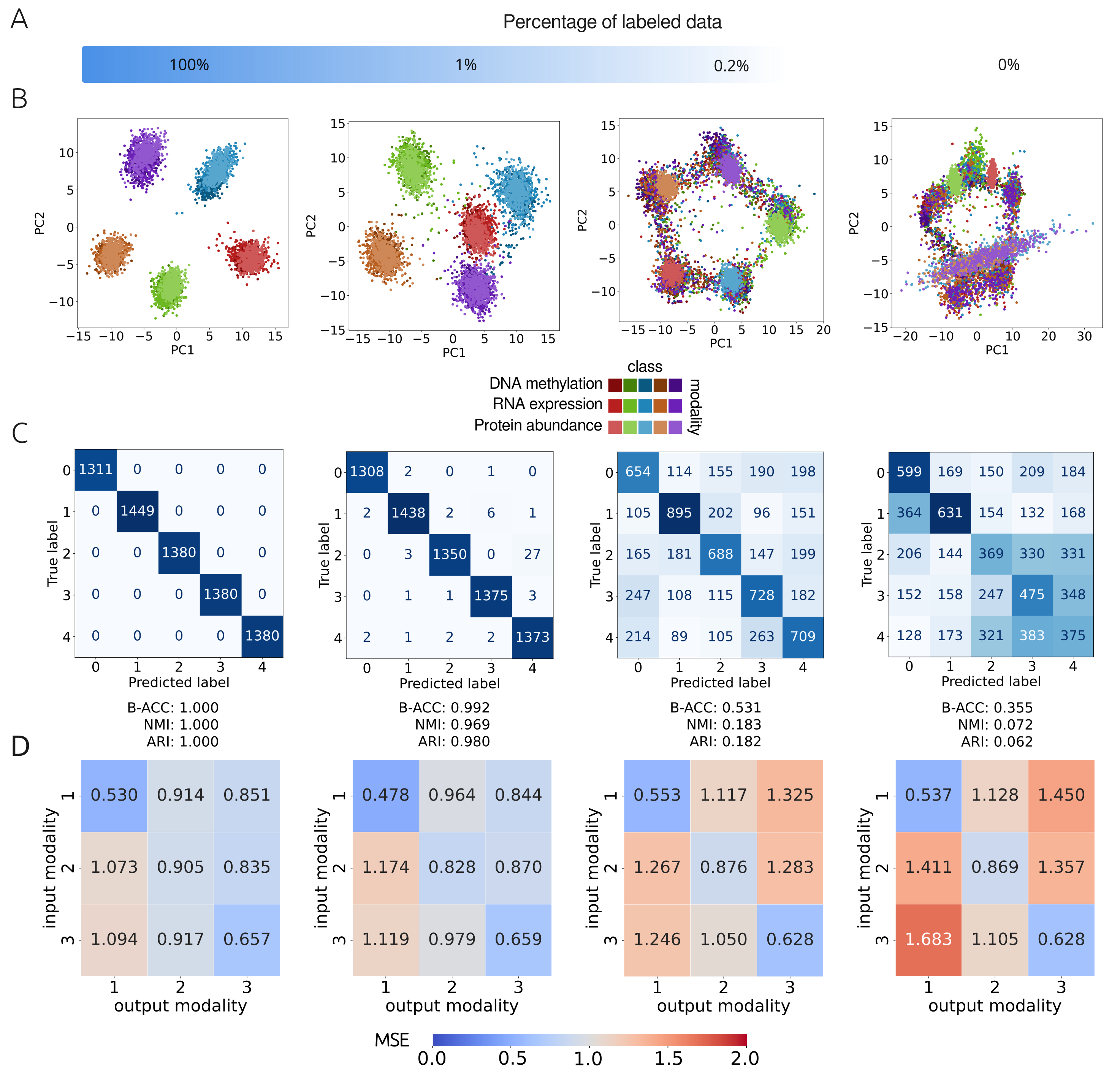}
    \caption{Latent Alignment obtained with MODIS - INTERSIM dataset with cluster mean shift parameter d = 0.5}
    \label{intersim_0.5d}
\end{figure}

\end{document}